\pdfoutput=1
\documentclass[10pt,twocolumn,letterpaper]{article}

\usepackage[letterpaper,top=18mm,bottom=19mm,left=16mm,right=16mm,columnsep=6.2mm]{geometry}
\usepackage{mathptmx}
\usepackage[T1]{fontenc}
\usepackage{microtype}
\usepackage{amsmath,amssymb,amsthm}
\usepackage{mathtools}
\usepackage{graphicx}
\usepackage{booktabs}
\usepackage{tabularx}
\usepackage[table]{xcolor}
\usepackage{caption}
\usepackage{hyperref}

\definecolor{iccvblue}{RGB}{24,72,138}
\definecolor{gainpos}{RGB}{15,118,110}
\definecolor{gainneg}{RGB}{180,83,9}
\definecolor{deprinttint}{RGB}{242,246,252}
\definecolor{venuetext}{RGB}{100,116,139}
\definecolor{panelbg}{RGB}{246,248,252}

\hypersetup{
  colorlinks=true,
  linkcolor=iccvblue,
  citecolor=iccvblue,
  urlcolor=iccvblue
}

\newtheorem{proposition}{Proposition}
\newtheorem{corollary}{Corollary}

\newcommand{\R}{\mathbb{R}}
\newcommand{\norm}[1]{\left\|#1\right\|}
\newcommand{\pos}[1]{\textcolor{gainpos}{\textbf{#1}}}
\newcommand{\negv}[1]{\textcolor{gainneg}{#1}}
\newcommand{\venue}[1]{\textcolor{venuetext}{\textsf{\scriptsize #1}}}
\newcommand{\vcSix}[1]{\raisebox{-0.65\normalbaselineskip}[0pt][0pt]{#1}}
\newcommand{\cc}{\cellcolor{deprinttint}}

\begin{document}

% =========================================================================
% Clean Page-1 Title + Abstract + Intro (floats start cleanly at top of Page 2)
\title{\Large\bfseries DePrint: Removing Intrinsic Error Fingerprints\\Across Visual Anomaly Detectors}
\author{Ning Zhu\\[2pt]\normalsize Xi'an University of Technology}
\date{}
\maketitle

% =========================================================================
\begin{abstract}
Visual anomaly detectors span memory-bank retrieval, mathematical modeling, and feature reconstruction, yet nearly all share one scoring bottleneck: they collapse a $D$-dimensional patch residual $\Delta = q - \hat{q} \in \R^{D}$ into a pointwise scalar norm $\|\Delta\|_2$.
This reduction implicitly treats normal residuals as isotropic, spatially independent noise.
We show that this assumption fails systematically: every detector paired with a finite normal support set leaves a reproducible \emph{intrinsic residual fingerprint} governed by two incompatible geometries.
In the channel domain, benign shifts concentrate inside a detector-specific low-rank linear subspace, while true defects inject orthogonal energy.
In the local spatial domain, $3\times 3$ residual magnitudes form a non-convex order ideal uniquely determined by its sparse Pareto-maximal frontier.
Canceling both domains obeys a strict causal order: spatial calibration on raw scores degrades accuracy because unremoved channel drift inflates the upper normal boundary, whereas removing channel drift first tightens the frontier.
We introduce \textbf{DePrint}, a training-free, model-agnostic calibrator combining a rank-$32$ uncentered SVD basis (Stage~1, \texttt{DePrint-S1}) with $32$ greedy spatial Pareto prototypes (Stage~2, \texttt{DePrint-S1+S2}) in $48.6$\,KB per class and $2.69$\,ms per image.
Controlled $2\times 2$ diagnostics on VisA confirm strong channel anisotropy ($r_{\mathrm{eff}}/D = 0.159$--$0.417$ vs.\ $0.957$ null), diagonal dominance in causal transfer (up to $+14.35$\,pp on PaDiM), and collapse of convex 9-D Mahalanobis modeling.
Across 27-class, 3-seed benchmarks spanning seven few-shot detectors, \texttt{DePrint-S1} improves the 6-metric mean on all 14 detector-dataset pairs (up to $+7.44$ I-AUC on PaDiM and $+0.69$ Avg.\ on official 2/4-shot INP-Former with zero retraining).
On saturated full-shot multi-class checkpoints (Dinomaly and INP-Former), DePrint remains near-neutral overall while lifting VisA pixel localization via Stage~1 and raising INP-Former VisA I-AUC to $99.11\%$ via Stage~2.
\end{abstract}

% =========================================================================
\section{Introduction}

\begin{figure*}[t]
\centering
\includegraphics[width=\linewidth]{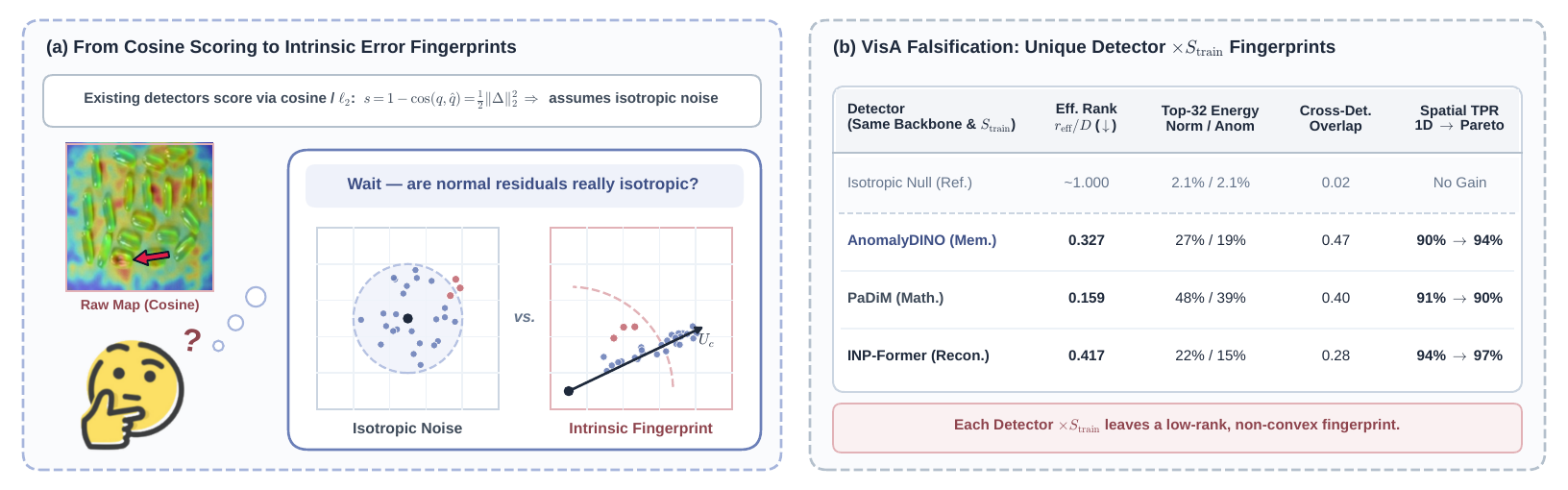}
\caption{\textbf{Every visual anomaly detector leaves an intrinsic error fingerprint rather than isotropic noise.} (a) Existing detectors score feature residuals via cosine or $\ell_2$ distance ($s = 1 - \cos(q,\hat{q}) = \frac{1}{2}\|\Delta\|_2^2$), implicitly assuming isotropic noise; a real raw cosine map on VisA \texttt{capsules} exhibits widespread false alarms on normal capsules outside the true defect (red arrow). (b) Controlled falsification on VisA under a shared \texttt{DINOv2-reg-B/14} ($R448\text{--}C392$) backbone and identical 4-shot support $S_{\mathrm{train}}$ across three paradigms (AnomalyDINO, PaDiM, INP-Former) reveals strong low-rank channel anisotropy, low cross-detector overlap, and non-convex spatial structure unique to $\text{Detector} \times S_{\mathrm{train}}$.}
\label{fig:teaser}
\end{figure*}

Visual anomaly detection aims to identify and localize unseen defects using only defect-free reference images.
Prior methods span four primary families: feature reconstruction (CNN reverse distillation such as RD4AD~\cite{deng2022rd4ad}, and Transformer decoders such as Dinomaly~\cite{guo2025dinomaly} and INP-Former~\cite{luo2025inpformer}), memory-bank retrieval (PatchCore~\cite{roth2022patchcore}, AnomalyDINO~\cite{damm2025anomalydino}), mathematical modeling (PaDiM~\cite{defard2021padim}, FastRecon~\cite{fang2023fastrecon}, SubspaceAD~\cite{subspacead2025}), and vision-language models (WinCLIP~\cite{jeong2023winclip}, AnomalyCLIP~\cite{zhou2024anomalyclip}). Visual-residual detectors across the first three paradigms construct a $D$-dimensional feature residual $\Delta(x) = q(x) - \hat{q}_{f,\mathcal{S}}(x) \in \R^{D}$ against a normal support set $\mathcal{S} = \{I_1, \dots, I_K\}$; pure text-prompted VLMs instead score scalar visual-text similarities, yet still form cross-modal residuals $q(x) - t_{\mathrm{norm}}$ (analyzed separately in Appendix~\ref{app:vlm}).
Although these detectors construct $\hat{q}_{f,\mathcal{S}}(x)$ in very different ways, nearly all terminate at the same bottleneck: they collapse $\Delta(x)$ directly into a pointwise scalar norm $s(x) = \|\Delta(x)\|_2$.

Normal residuals are not isotropic noise.
Every detector $f$ paired with a finite normal support set $\mathcal{S}$ produces a reproducible, structured error pattern on normal test images, which we term its \emph{intrinsic residual fingerprint} (Fig.~\ref{fig:teaser}).
This fingerprint stems from two coupled sources: \emph{model-induced bias} (patch-grid phase sensitivity in Vision Transformers~\cite{dosovitskiy2021vit,oquab2024dinov2,darcet2024registers}, convolutional boundary responses in CNNs, diagonal whitening in PaDiM, ridge shrinkage in FastRecon, or low-pass smoothing in reconstruction decoders) and \emph{support-induced coverage bias} (benign pose, lighting, or texture-phase variations absent from the $K$ normal references).
When scoring reduces $\Delta$ directly to $\|\Delta\|_2$, the detector penalizes every test image for its own predictable fingerprint and buries subtle anomalies beneath structured normal false alarms.

Inspecting residual fields across paradigms shows why no single isotropic metric or linear operator can remove this fingerprint: its channel and spatial components obey incompatible geometries.
In the channel domain, first-order nuisance shifts confine normal residuals to a low-rank linear subspace $U \in \R^{D\times r}$ anchored at the exact-match origin $\Delta = \mathbf{0}$, while true defects project into the orthogonal complement $U^\perp$.
In the local spatial domain, residual magnitudes over $3\times 3$ neighborhoods follow coordinate-wise dominance: a test block is normal whenever every coordinate falls below some observed normal pattern, forming a \emph{non-convex order ideal} $\bigcup_{p\in\mathcal{P}}[0,p] \subset \R_{\ge 0}^{9}$.
Approximating this region with symmetric $\ell_2$ balls or a convex 9-D Mahalanobis ellipsoid fails sharply because symmetric norms penalize clean low-score regions and convex envelopes swallow interior defects.

Moreover, canceling the two domains obeys a strict causal ordering.
Applying spatial Pareto calibration directly to raw scores degrades accuracy below the uncalibrated baseline because unremoved channel drift inflates the upper normal boundary and masks true anomalies.
Projecting out the channel drift subspace first, then scoring one-sided excess against a compact spatial Pareto frontier, resolves both failure modes together.

We instantiate this mechanism as \textbf{DePrint} (Fig.~\ref{fig:overview}), a training-free, model-agnostic calibrator requiring only $48.6$\,KB per category (float16) and $2.69$\,ms per image on GPU.
Unlike prior post-processing heuristics that apply global scaling or isotropic Gaussian smoothing to scalar score maps, DePrint operates directly on the residual vector field with provable zero-loss Pareto pruning and conservative $1$-Lipschitz compression bounds.
Our contributions are:
\begin{enumerate}
  \setlength{\itemsep}{2pt}
  \item \textbf{Geometric theory and $2\times 2$ falsification:} We formalize the incompatible channel and spatial geometries of detector-support residual fingerprints (Props.~\ref{prop:svd}--\ref{prop:compress}) and validate them via controlled $2\times 2$ diagnostics under a shared backbone (both PaDiM and SubspaceAD variants) that show strong channel anisotropy, detector-specific causal transfer, and spatial non-convexity.
  \item \textbf{Cascaded fingerprint cancellation (DePrint):} We introduce a $48.6$\,KB training-free calibrator combining origin-anchored rank-$32$ channel drift removal (\texttt{DePrint-S1}) with a $32$-prototype one-sided spatial Pareto bank (\texttt{DePrint-S1+S2}), and demonstrate why channel removal must precede spatial modeling.
  \item \textbf{Cross-paradigm benchmark and design ablations:} Across 27-class, 3-seed benchmarks covering seven few-shot detectors (42/42 positive VisA entries) and two official full-shot multi-class SOTA checkpoints, together with five ablations and robustness/VLM stress tests, we quantify both DePrint's broad few-shot gains and its transparent operating boundaries.
\end{enumerate}

% =========================================================================
\section{Related Work}

\noindent\textbf{Feature Reconstruction and Anomaly Synthesis.} Reconstruction detectors score anomalies via discrepancies between frozen encoder features and learned decoders. Reverse-distillation frameworks reconstruct multi-scale CNN features (RD4AD~\cite{deng2022rd4ad}, RD++~\cite{tien2023rdpp}, DeSTSeg~\cite{zhang2023destseg}, EfficientAD~\cite{batzner2024efficientad}), while unified multi-class architectures address cross-category identity shortcuts via masked Transformer decoders (UniAD~\cite{you2022uniad}), state-space models (MambaAD~\cite{he2024mambaad}), linear-attention decoders (Dinomaly~\cite{guo2025dinomaly}), and intrinsic normal prototypes (INP-Former~\cite{luo2025inpformer}). Complementary synthesis-based methods train discriminative heads or dual-subspace reconstructors on pseudo-defects in image or feature space, including CutPaste~\cite{li2021cutpaste}, DR{\AE}M~\cite{zavrtanik2021draem}, NSA~\cite{schluter2022nsa}, DSR~\cite{zavrtanik2022dsr}, SimpleNet~\cite{liu2023simplenet}, RealNet~\cite{zhang2024realnet}, and GLASS~\cite{chen2024glass}. Even after multi-class training, inference collapses multi-layer residuals into pointwise cosine distances, leaving category-specific normal drift uncalibrated.

\noindent\textbf{Memory-Bank, Mathematical, and Flow Modeling.} Memory-bank methods compare query patches against stored normal patch banks via nearest-neighbor search (SPADE~\cite{cohen2020spade}, PatchCore~\cite{roth2022patchcore}, noise-robust SoftPatch~\cite{jiang2022softpatch}, and DINOv2-based AnomalyDINO~\cite{damm2025anomalydino}). Mathematical modeling approaches fit explicit statistical, geometric, or likelihood models over normal patch features, including per-location Gaussians (PaDiM~\cite{defard2021padim}), closed-form ridge regression (FastRecon~\cite{fang2023fastrecon}), PCA subspaces (SubspaceAD~\cite{subspacead2025}), and normalizing-flow density estimators (CFLOW-AD~\cite{gudovskiy2022cflow}, CS-Flow~\cite{rudolph2022csflow}, MSFlow~\cite{zhou2025msflow}). Across all these visual-residual detectors, the final scoring step discards residual vector direction and local spatial partial-order structure.

\noindent\textbf{Few-Shot and Vision-Language Anomaly Detection.} Few-shot industrial inspection mitigates scarce normal references through feature registration (RegAD~\cite{huang2022regad}), graph-based patch aggregation (GraphCore~\cite{xie2023graphcore}), or unlabeled mutual scoring (MuSc~\cite{li2024musc}). Vision-language detectors leverage CLIP~\cite{radford2021clip} and multimodal LLMs via textual prompts or in-context residual prompting (WinCLIP~\cite{jeong2023winclip}, AnomalyCLIP~\cite{zhou2024anomalyclip}, PromptAD~\cite{li2024promptad}, InCTRL~\cite{zhu2024inctrl}, AnomalyGPT~\cite{gu2024anomalygpt}). While pure text-prompted VLMs score patches via scalar visual-text dot products rather than a visual reference $\hat{q}_{f,\mathcal{S}}$, we also evaluate post-hoc DePrint calibration on official zero-shot AnomalyCLIP's cross-modal residuals $q - t_{\mathrm{norm}}$ in Appendix~\ref{app:vlm}.

\begin{table*}[t]
\centering
\renewcommand{\arraystretch}{1.13}
\setlength{\tabcolsep}{4.6pt}
\caption{\textbf{Controlled $2\times 2$ fingerprint diagnostic on VisA-12 (4-shot, shared \texttt{DINOv2-reg-B/14} at $R448\text{--}C392$, $r=32$, $K_p=32$).} Note: This table uses shared \texttt{DINOv2-reg-B/14} to match INP-Former's encoder, whereas standalone AnomalyDINO runs (Table~\ref{tab:main_fewshot}, ablations, and Suite~F) use non-register \texttt{DINOv2-B/14} ($448\times 448$). \textbf{Panel A:} Channel effective rank $r_{\mathrm{eff}}/D$, Top-32 residual energy (normal / anomaly), and cross-class $U_c$-projection transfer ($\Delta$ I-AUC, pp). \textbf{Panel B:} Cross-detector Grassmann overlap between top-32 drift bases $U_c$ (Prop.~\ref{prop:svd}; diagonal = disjoint splits of the same detector) and causal $U_c$-transfer ($\Delta$ I-AUC when projecting row detector with column detector's $U_c$). \textbf{Panel C:} Spatial 9-D non-convexity (TPR at 5\% normal FPR and 9-D effective rank $r_{\mathrm{eff},9}$).}
\label{tab:diagnostic_2x2}
\resizebox{\linewidth}{!}{%
\begin{tabular}{@{}lccc@{\hspace{9pt}}ccc@{\hspace{9pt}}ccc@{\hspace{9pt}}cccc@{}}
\toprule
 & \multicolumn{3}{c}{\textbf{Panel A: Channel Anisotropy}} & \multicolumn{3}{c}{\textbf{Panel B1: Grassmann Overlap}} & \multicolumn{3}{c}{\textbf{Panel B2: $U_c$ Transfer ($\Delta$ I-AUC)}} & \multicolumn{4}{c}{\textbf{Panel C: Spatial TPR@5\%FPR}} \\
\cmidrule(lr){2-4}\cmidrule(lr){5-7}\cmidrule(lr){8-10}\cmidrule(lr){11-14}
\textbf{Detector} & $r_{\mathrm{eff}}/D$ & \textbf{$E_{32}$ (N/A)} & \textbf{X-Cls} & \textbf{ADINO} & \textbf{PaDiM} & \textbf{INP-F} & \textbf{ADINO} & \textbf{PaDiM} & \textbf{INP-F} & \textbf{1D} & \textbf{Maha-9} & \textbf{Pareto} & $r_{\mathrm{eff},9}$ \\
\midrule
Isotropic Null & 0.957 & 2.1\% / 2.1\% & --- & 0.02 & 0.02 & 0.02 & --- & --- & --- & --- & --- & --- & 0.999 \\
AnomalyDINO & \textbf{0.327} & 27.2\% / 19.0\% & +0.67 & \cc\textbf{0.80} & 0.59 & 0.36 & \cc\pos{+0.79} & +0.49 & +0.53 & 89.5\% & 66.7\% & \textbf{93.6\%} & 0.131 \\
PaDiM       & \textbf{0.159} & 47.5\% / 38.7\% & +3.04 & 0.59 & \cc\textbf{0.85} & 0.21 & +2.64 & \cc\pos{+14.35} & +0.18 & \textbf{90.7\%} & 35.2\% & 90.3\% & 0.128 \\
INP-Former  & \textbf{0.417} & 22.0\% / 15.3\% & +0.27 & 0.36 & 0.21 & \cc\textbf{0.77} & +0.26 & +0.25 & \cc\pos{+0.41} & 94.1\% & 91.1\% & \textbf{97.1\%} & 0.126 \\
\bottomrule
\end{tabular}}
\end{table*}

\noindent\textbf{Post-Hoc Calibration and Subspace OOD Scoring.} In image-level out-of-distribution (OOD) detection, post-hoc methods calibrate frozen classifiers using class-conditional Mahalanobis distances~\cite{lee2018mahalanobis}, activation rectification (ReAct~\cite{sun2021react}), or residual norms orthogonal to a principal feature subspace (ViM~\cite{wang2022vim}). In contrast, industrial anomaly post-processing still relies mainly on global quantile scaling or isotropic Gaussian smoothing of scalar score maps. DePrint brings geometric post-hoc calibration to dense patch residual fields, decoupling origin-anchored low-rank channel drift from non-convex $3\times 3$ spatial Pareto order ideals without retraining the host detector.

% =========================================================================
\section{Geometry of Intrinsic Residual Fingerprints}

Let a frozen host detector $f$ fitted on support set $\mathcal{S} = \{I_1, \dots, I_K\}$ map any image $I$ to a $D$-dimensional patch residual field $\Delta(I) \in \R^{(HW)\times D}$ with rows $\Delta_u = q_u - \hat{q}_u$.

\subsection{Channel Fingerprint: Low-Rank Linear Drift}

\begin{figure*}[t]
\centering
\includegraphics[width=\linewidth]{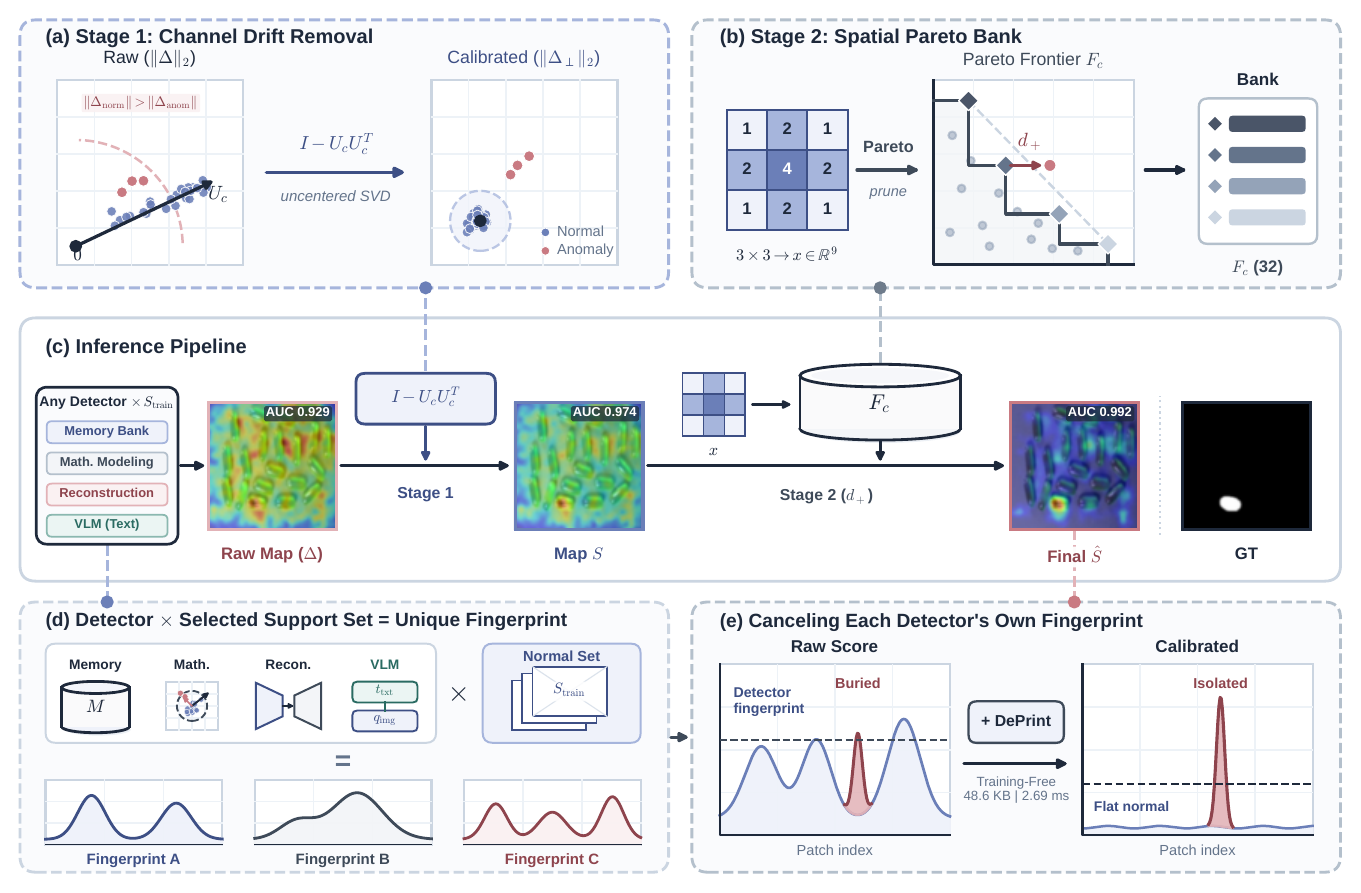}
\caption{\textbf{Overview of DePrint.} (a) \textbf{Stage 1 (Channel Drift Removal):} Raw normal residuals along low-rank drift basis $U_c$ can exceed anomalous norms ($\|\Delta_{\mathrm{norm}}\| > \|\Delta_{\mathrm{anom}}\|$); projecting out $U_c$ via $I - U_c U_c^\top$ (uncentered SVD, Prop.~\ref{prop:svd}) collapses normal drift toward the origin while preserving orthogonal defect residuals $\Delta_\perp$. (b) \textbf{Stage 2 (Spatial Pareto Bank):} Gaussian-weighted $3\times 3$ blocks $x \in \mathbb{R}_{\ge 0}^9$ are pruned to the non-convex Pareto frontier and compressed into $K_p=32$ prototypes $F_c$ (Props.~\ref{prop:pareto}--\ref{prop:compress}), scoring defects via one-sided excess $d_+$. (c) \textbf{Inference Pipeline:} Real progression on VisA \texttt{capsules} (\texttt{bad/065}): Stage~1 suppresses broad normal false alarms ($0.929\to 0.974$), and Stage~2 isolates the true defect matching GT ($0.974\to 0.992$). (d--e) Every $\text{Detector} \times S_{\mathrm{train}}$ pair leaves a unique fingerprint canceled in $2.69$\,ms without retraining.}
\label{fig:overview}
\end{figure*}

When a normal query patch differs from its matched reference by a small benign nuisance perturbation (Fig.~\ref{fig:overview}(a)) $\delta\theta \in \R^{p}$ ($p \ll D$, e.g., sub-pixel grid phase shift, slight pose change, or illumination variation), first-order linearization yields
\begin{equation}
\Delta_u = q_u - \hat{q}_u \approx J_f(\hat{q}_u)\,\delta\theta + \varepsilon_u.
\label{eq:taylor}
\end{equation}
Because $\Delta = \mathbf{0}$ corresponds to exact physical agreement between query and normal reference, any non-zero mean drift $\bar{\Delta} \neq \mathbf{0}$ induced by $\mathcal{S}$ is itself benign bias relative to the origin.

\begin{proposition}[Optimal Origin-Anchored Drift Cancellation]
\label{prop:svd}
For category $c$, let $R_c \in \R^{N_{\mathrm{fg}}\times D}$ stack out-of-sample foreground normal residuals with thin SVD $R_c = W\Sigma V^\top$ ($\sigma_1 \ge \dots \ge \sigma_D \ge 0$). Among all rank-$r$ orthogonal projectors $P = UU^\top$ ($U\in\R^{D\times r}$, $U^\top U = I_r$), taking the drift basis $U_c = V_{:,1:r}$ uniquely minimizes average squared normal residual energy under orthogonal projection $\Delta_\perp = \Delta(I - U_c U_c^\top)$:
\begin{equation}
\min_{U^\top U = I_r} \frac{1}{N_{\mathrm{fg}}}\norm{R_c - R_c UU^\top}_F^2 = \frac{1}{N_{\mathrm{fg}}}\sum_{j=r+1}^{D} \sigma_j^2,
\end{equation}
whereas mean-centered PCA minimizes variance around the sample mean $\bar{\Delta}$ and leaves a theoretical mean-offset floor $\|\bar{\Delta}_\perp\|_2^2$ whenever $\bar{\Delta}$ has energy orthogonal to the centered top-$r$ covariance subspace.
\end{proposition}
\begin{proof}
By the Eckart--Young--Mirsky theorem~\cite{eckart1936svd}, $R_c U_c U_c^\top$ is the optimal rank-$r$ Frobenius approximation to $R_c$, directly minimizing $\|R_c(I-UU^\top)\|_F^2 = \sum_{j=r+1}^D \sigma_j^2$. Centering around $\bar{\Delta}$ optimizes variance around $\bar{\Delta}$ rather than energy relative to $\mathbf{0}$, leaving $\bar{\Delta}(I - U_{\mathrm{pca}}U_{\mathrm{pca}}^\top)$ uncanceled.
\end{proof}

\noindent\textbf{Cross-set stability and empirical centering.} On VisA under a shared \texttt{DINOv2-reg-B/14} ($R448\text{--}C392$) backbone, the Grassmann overlap $\frac{1}{r}\|U_A^\top U_B\|_F^2$ between top-$32$ bases fitted on disjoint 4-shot support splits reaches $0.80$ (AnomalyDINO), $0.85$ (PaDiM), and $0.77$ (INP-Former), versus $0.02$ under isotropic noise (Sec.~\ref{sec:diag}). In practice, on $\ell_2$-normalized pretrained visual features, patch residuals aggregated across spatial positions have a small global mean $\|\bar{\Delta}\|_2$ relative to their directional variance, and its residual component largely aligns with the leading singular directions; consequently $\|\bar{\Delta}_\perp\|_2^2$ is negligible and uncentered SVD matches centered PCA within $<0.02$\,pp on both MVTec-AD and VisA (Suite~B). We therefore adopt uncentered SVD for its origin-anchored formulation ($\Delta=\mathbf{0}$) and simpler parameterization (no separate mean vector) rather than claiming an empirical accuracy margin over centered PCA.

\subsection{Spatial Fingerprint: Non-Convex Pareto Order Ideal}

After orthogonal projection removes channel drift (Fig.~\ref{fig:overview}(a)), residual errors remain spatially structured across local neighborhoods (Fig.~\ref{fig:overview}(b)). Let $S^{(i)} = \|\Delta_\perp^{(i)}\|_2 / q_{95} \in \R_{\ge 0}^{H\times W}$ be the channel-calibrated normal residual map. Unfolding Gaussian-weighted $3\times 3$ windows yields normal blocks $E = \{p_1,\dots,p_M\} \subset \R_{\ge 0}^{9}$. Under the coordinate-wise partial order $\preceq$, the one-sided excess distance
\begin{equation}
d_+(x, E) = \min_{p \in E} \norm{\max(x - p,\, \mathbf{0})}_2
\label{eq:onesided}
\end{equation}
induces the zero-excess normal region $\mathcal{N}(E) = \bigcup_{p\in E}[\mathbf{0}, p]$, a \emph{non-convex order ideal} in $(\R_{\ge 0}^9, \preceq)$.

\begin{proposition}[Exact Pareto Fingerprint Reduction]
\label{prop:pareto}
Let $\mathcal{P}(E) = \{p\in E : \nexists\, p'\in E,\ p'\succeq p,\ p'\neq p\}$ be the Pareto-maximal frontier of finite $E\subset\R_{\ge 0}^9$. Then for every $x \in \R_{\ge 0}^9$, $d_+(x, E) = d_+(x, \mathcal{P}(E))$.
\end{proposition}
\begin{proof}
Since $\mathcal{P}(E)\subseteq E$, $d_+(x,E)\le d_+(x,\mathcal{P}(E))$. Conversely, for any $p\in E$, finiteness and transitivity of $\succeq$ guarantee $p^*\in\mathcal{P}(E)$ with $p^*\succeq p$, so $0\le\max(x_j-p^*_j,0)\le\max(x_j-p_j,0)$ for all $j$, yielding $d_+(x,\mathcal{P}(E))\le d_+(x,E)$.
\end{proof}

\begin{corollary}[Failure of Convex and Symmetric Approximations]
\label{cor:nonconvex}
Replacing $\mathcal{N}(E)$ by a convex envelope strictly enlarges the normal region whenever $\mathcal{P}(E)$ contains incomparable boundary patterns ($p^{(1)} \not\preceq p^{(2)}$ and $p^{(2)} \not\preceq p^{(1)}$), swallowing interior defects; symmetric $\ell_2$ distance $\|x-p\|_2$ violates order monotonicity ($x\preceq p \not\Rightarrow \|x-p\|_2=0$).
\end{corollary}

\begin{proposition}[Conservative Coverage and $1$-Lipschitz Stability]
\label{prop:compress}
For any compressed category prototype bank $F_c \subseteq \mathcal{P}(E)$ with one-sided covering radius $\epsilon = \max_{p\in\mathcal{P}(E)} d_+(p, F_c)$, every $x, y \in \R_{\ge 0}^{9}$ satisfies
\begin{equation}
\begin{gathered}
0 \le d_+(x, F_c) - d_+(x, \mathcal{P}(E)) \le \epsilon, \\[2pt]
|d_+(x, F_c) - d_+(y, F_c)| \le \norm{x - y}_2.
\end{gathered}
\end{equation}
\end{proposition}
\begin{proof}
Since $F_c\subseteq\mathcal{P}(E)$, $d_+(x,F_c)\ge d_+(x,\mathcal{P}(E))$ holds immediately (preventing false-negative shrinkage). For any $p\in\mathcal{P}(E)$ and its nearest $q\in F_c$, $\max(x-q,0) \le \max(x-p,0) + \max(p-q,0)$ entrywise, so the triangle inequality gives $d_+(x,F_c)\le \|\max(x-p,0)\|_2 + d_+(p,F_c) \le d_+(x,\mathcal{P}(E)) + \epsilon$. Entrywise $|\max(x-q,0)-\max(y-q,0)| \le |x-y|$ establishes $1$-Lipschitz continuity.
\end{proof}

\subsection{Empirical Verification: Controlled $2\times 2$ Diagnostic}

\label{sec:diag}

To directly test whether normal residuals behave as isotropic noise or detector-specific fingerprints, we run a strictly controlled $2\times 2$ diagnostic on VisA (12 categories, 4-shot, $r=32$, $K_p=32$) where all three paradigms share the \emph{exact same frozen} \texttt{DINOv2-reg-B/14}~\cite{oquab2024dinov2,darcet2024registers} ($R448\text{--}C392$) encoder and identical support splits: \textbf{AnomalyDINO} (Memory), \textbf{PaDiM} (Math., per-position diagonal Gaussian whitening with $\epsilon=10^{-3}$), and \textbf{INP-Former} (Recon.). Table~\ref{tab:diagnostic_2x2} reports the full results.

Four findings emerge directly from Table~\ref{tab:diagnostic_2x2}:
\begin{enumerate}
  \setlength{\itemsep}{2pt}
  \item \textbf{Strong channel anisotropy (Panel A):} All three detectors collapse far below the isotropic null ($r_{\mathrm{eff}}/D = 0.954$--$0.959$). PaDiM exhibits the sharpest directional concentration ($r_{\mathrm{eff}}/D = 0.159$, Top-32 normal energy $47.5\%$ vs.\ $2.1\%$ null), followed by AnomalyDINO ($0.327$, $27.2\%$) and INP-Former ($0.417$, $22.0\%$). Crucially, normal top-32 energy consistently exceeds anomaly energy ($47.5\% > 38.7\%$, $27.2\% > 19.0\%$, $22.0\% > 15.3\%$), explaining why orthogonal projection suppresses normal drift more than true defects.
  \item \textbf{Detector- and support-specific subspaces (Panel B):} Self-consistency across disjoint support splits is high ($0.77$--$0.85$), while cross-detector Grassmann overlap drops sharply ($0.21$--$0.59$). In causal $U_c$-projection transfer, every row peaks strictly on its own diagonal basis: PaDiM gains $\mathbf{+14.35}$\,pp using its own basis versus at most $+2.64$\,pp from AnomalyDINO or $+0.18$\,pp from INP-Former; conversely, transferring PaDiM's basis to AnomalyDINO or INP-Former yields only $+0.49$ and $+0.25$\,pp (below their own $+0.79$ and $+0.41$\,pp). Under cross-class support perturbation (Col.\ \textbf{X-Cls}), PaDiM drops from $+14.35$\,pp to $+3.04$\,pp (AnomalyDINO $+0.79\to +0.67$; INP-Former $+0.41\to +0.27$).
  \item \textbf{Severe spatial non-convexity with detector-dependent Pareto behavior (Panel C):} High-score normal $3\times 3$ blocks are strongly low-dimensional ($r_{\mathrm{eff},9} = 0.126$--$0.131$ vs.\ $0.999$ null), and convex 9-D Mahalanobis modeling collapses across every detector: $35.2\%$ on PaDiM, $66.7\%$ on AnomalyDINO, and $91.1\%$ on INP-Former. One-sided Pareto lifts TPR@5\%FPR over 1-D scalars on AnomalyDINO ($89.5\%\to 93.6\%$) and INP-Former ($94.1\%\to 97.1\%$). On PaDiM, however, 1-D scoring is already strong ($90.7\%$) and Pareto ($90.3\%$) does not exceed it, matching PaDiM's channel-projection dominance in Sec.~\ref{sec:exp}.
  
  \item \textbf{Negative Pareto-bank P-$F_1$ transfer (Panel D) and SubspaceAD cross-check:} Across all nine entries of the cross-detector Pareto-bank ($F_c$) $\Delta$ P-$F_1$ matrix, transferring spatial prototypes degrades P-$F_1$ ($-0.54$ to $-2.37$\,pp; PaDiM self-transfer $-1.39$\,pp, AnomalyDINO $-1.51$\,pp, INP-Former $-2.03$\,pp), showing that spatial banks must be fitted per detector/support and confirming that hard-floor Pareto scoring trades P-$F_1$ for I-AUC/PRO. Replacing PaDiM with \textbf{SubspaceAD} under the identical protocol yields the same directional conclusions ($r_{\mathrm{eff}}/D = 0.367$, cross-detector overlap $0.01$, diagonal transfer peaks $+0.79 / +0.011 / +0.41$\,pp, and convex Maha-9 collapsing to $76.1\%$ vs.\ Pareto $97.1\%$), confirming that our $2\times 2$ findings do not depend on a specific mathematical detector.
\end{enumerate}

% =========================================================================
\section{Method: DePrint}

Figure~\ref{fig:overview} illustrates the two-stage pipeline of \textbf{DePrint}.

\noindent\textbf{Stage 1: Channel Drift Removal (\texttt{DePrint-S1}).} Given out-of-sample normal residual vectors $R_c \in \R^{N_{\mathrm{fg}}\times D}$ for category $c$, we compute the top-$r$ ($r=32$) right singular vectors $U_c \in \R^{D\times 32}$ via uncentered thin SVD. At inference, each patch residual $\Delta_u$ is orthogonally projected onto $U_c^\perp$:
\begin{equation}
s_{\mathrm{ch}}(\Delta_u) = \norm{\Delta_u - (\Delta_u U_c) U_c^\top}_2.
\end{equation}

\noindent\textbf{Stage 2: Spatial Pareto Bank (\texttt{DePrint-S1+S2}).} On the Stage-1 calibrated normal maps $S^{(i)} = s_{\mathrm{ch}}(\Delta^{(i)}) / q_{95,c}$ (where $q_{95,c}$ is the 95th percentile of normal foreground scores), we unfold $3\times 3$ neighborhoods weighted by the normalized binomial Gaussian kernel $\frac{1}{16}[1,2,1;2,4,2;1,2,1]$ (square-root weighted per coordinate) to form $E_c \subset \R_{\ge 0}^{9}$. We prune $E_c$ to its exact Pareto frontier $F_{0,c} = \mathcal{P}(E_c)$ (reducing $\sim 4{,}000$ blocks to $\sim 477$ frontier points on average) and greedily select $K_p = 32$ prototypes $F_c \in \R^{32\times 9}$ minimizing the $99\%$-quantile one-sided residual over $E_c$. At inference, each $3\times 3$ test block $x_u \in \R_{\ge 0}^9$ is scored via $\hat{S}_u = d_+(x_u, F_c)$ (Eq.~\ref{eq:onesided}).

\noindent\textbf{Unified Per-Category Calibration Protocol.} Even when the host detector uses a single shared multi-class decoder (Dinomaly, INP-Former), DePrint fits $(U_c, q_{95,c}, F_c)$ strictly per category to prevent cross-category boundary inflation. Normal calibration residuals are collected without extra data via:
(i) \textbf{Leave-One-Image-Out (LOIO)} for training-free detectors (AnomalyDINO, PatchCore, PaDiM, FastRecon, SubspaceAD);
(ii) \textbf{$K$-fold cross-validation} (LOIO for $K\le 4$) for few-shot trained detectors (RD4AD); and
(iii) \textbf{zero-retraining post-hoc calibration} on the exact normal training images for official released checkpoints: for 2/4-shot INP-Former, strictly its exact 2 or 4 training images; for full-shot multi-class Dinomaly and INP-Former, exact streaming \texttt{float64} full-dataset Gram accumulation for Stage~1 (zero sampling variance) and stratified sampling ($N_{\mathrm{cal}}=64$) for Stage~2.

\noindent\textbf{Complexity and Footprint.} Per category at $D=768$ with $r=32$ and $K_p=32$, DePrint stores $(768\times 32 + 32\times 9 + 1)$ parameters, totaling $\mathbf{48.6\,\text{KB}}$ in float16 ($99.5$\,KB in float32). Online inference requires $O(HW(Dr + 9K_p))$ operations, measured at $\mathbf{2.69\,\text{ms/image}}$ on GPU for $32\times 32$ grids (Suite~D, Appendix Table~\ref{tab:ablations}).

% =========================================================================
% Interleave Table 1 right before Section 5 so it appears cleanly at top of Sec 5 page

\section{Experiments}\label{sec:exp}

\subsection{Experimental Setup}
We evaluate on \textbf{MVTec-AD}~\cite{bergmann2019mvtec} (15 categories) and \textbf{VisA}~\cite{zou2022visa} (12 categories) across all six standard metrics: image-level AUROC (I-AUC), AP (I-AP), and $F_1$-max (I-$F_1$), plus pixel-level AUROC (P-AUC), $F_1$-max (P-$F_1$), and AU-PRO (PRO). Our seven benchmarked detectors span all three visual-residual paradigms (Memory, Math., Recon.; AnomalyDINO uses non-register \texttt{DINOv2-B/14} at $448\times 448$, whereas Table~\ref{tab:diagnostic_2x2} uses shared \texttt{DINOv2-reg-B/14} at $R448\text{--}C392$); pure-text zero-shot VLM calibration (official AnomalyCLIP) is evaluated separately in Appendix~\ref{app:vlm} to keep few-shot and zero-shot protocols distinct. Unless stated otherwise, all few-shot benchmarks report unified 2/4-shot averages over 3 random seeds (\texttt{42/43/44}, 6 runs per detector-dataset pair). Additional 8-shot scaling and rotation-augmented calibration results are deferred to the Appendix; partial RD4AD full-shot runs are excluded from main tables.

% Place Table 2 cleanly on its own page right beside Section 5.3
\begin{table*}[p]
\centering
\renewcommand{\arraystretch}{1.07}
\setlength{\tabcolsep}{4.6pt}
\caption{\textbf{Few-shot cross-paradigm benchmark across seven detectors on MVTec-AD and VisA (\%).} All seven detectors report unified 2/4-shot averages across 3 seeds (\texttt{42/43/44}); AnomalyDINO uses non-register \texttt{DINOv2-B/14} ($448\times 448$); official INP-Former checkpoints (\texttt{DINOv2-reg-B/14}) are calibrated strictly on their exact 2/4 training images with zero extra data and zero retraining. Best per detector in \textbf{bold}; $|\Delta|\le 0.05$ shown in neutral black.}
\label{tab:main_fewshot}
\resizebox{\linewidth}{!}{%
\begin{tabular}{@{}ccclccccccccc@{}}
\toprule
 & & & & \multicolumn{3}{c}{\textbf{Image-Level}} & \multicolumn{3}{c}{\textbf{Pixel-Level}} & \multicolumn{3}{c}{\textbf{Gain over Baseline}} \\
\cmidrule(lr){5-7}\cmidrule(lr){8-10}\cmidrule(lr){11-13}
\textbf{Paradigm} & \textbf{Detector} & \textbf{Venue} & \textbf{Variant} & \textbf{I-AUC} & \textbf{I-AP} & \textbf{I-$F_1$} & \textbf{P-AUC} & \textbf{P-$F_1$} & \textbf{PRO} & \textbf{$\Delta$ I-AUC} & \textbf{$\Delta$ P-AUC} & \textbf{$\Delta$ Avg.} \\
\midrule
\rowcolor{panelbg}\multicolumn{13}{l}{\textbf{MVTec-AD} \quad \textcolor{venuetext}{\textsf{\scriptsize 15 categories, 2/4-shot 6-run mean}}} \\
\midrule
  & & & Baseline & 96.23 & 97.47 & 95.97 & 96.20 & 58.14 & 91.86 & --- & --- & --- \\
  & AnomalyDINO & \venue{WACV'25} & +DePrint-S1 & 96.66 & 97.89 & 96.36 & 96.82 & \textbf{59.16} & 92.79 & \pos{+0.43} & \pos{+0.62} & \pos{+0.63} \\
  \vcSix{Memory} & & & \cc\textbf{+DePrint-S1+S2} & \cc\textbf{97.23} & \cc\textbf{98.50} & \cc\textbf{97.12} & \cc\textbf{97.18} & \cc 58.20 & \cc\textbf{92.97} & \cc\pos{+1.00} & \cc\pos{+0.98} & \cc\pos{+0.89} \\
\cmidrule(lr){2-13}
  & & & Baseline & 85.68 & 93.23 & 92.23 & 93.75 & 50.27 & 81.20 & --- & --- & --- \\
  & PatchCore & \venue{CVPR'22} & +DePrint-S1 & 86.26 & 93.53 & 92.93 & \textbf{94.22} & \textbf{50.97} & \textbf{82.01} & \pos{+0.58} & \pos{+0.47} & \pos{+0.59} \\
  & & & \cc\textbf{+DePrint-S1+S2} & \cc\textbf{86.94} & \cc\textbf{94.00} & \cc\textbf{93.41} & \cc 94.06 & \cc 50.87 & \cc 81.41 & \cc\pos{+1.26} & \cc\pos{+0.31} & \cc\pos{+0.72} \\
\midrule
  & & & Baseline & 81.01 & 90.27 & 90.50 & 92.64 & 44.19 & 81.53 & --- & --- & --- \\
  & PaDiM & \venue{ICPR'21} & \textbf{+DePrint-S1} & 84.58 & 92.06 & 91.60 & \textbf{94.77} & \textbf{47.69} & \textbf{85.54} & \pos{+3.57} & \pos{+2.13} & \pos{+2.68} \\
  & & & \cc +DePrint-S1+S2 & \cc\textbf{85.33} & \cc\textbf{92.48} & \cc\textbf{91.75} & \cc 93.87 & \cc 46.56 & \cc 82.49 & \cc\pos{+4.32} & \cc\pos{+1.23} & \cc\pos{+2.06} \\
\cmidrule(lr){2-13}
  & & & Baseline & 85.68 & 93.15 & 92.39 & 94.13 & 51.57 & 82.23 & --- & --- & --- \\
  Math. & FastRecon & \venue{ICCV'23} & +DePrint-S1 & 86.23 & 93.41 & 92.66 & \textbf{94.15} & 51.50 & \textbf{82.33} & \pos{+0.55} & +0.02 & \pos{+0.19} \\
  & & & \cc\textbf{+DePrint-S1+S2} & \cc\textbf{87.10} & \cc\textbf{93.97} & \cc\textbf{93.35} & \cc 94.09 & \cc 51.36 & \cc 81.92 & \cc\pos{+1.42} & \cc $-$0.04 & \cc\pos{+0.44} \\
\cmidrule(lr){2-13}
  & & & Baseline & 94.34 & 97.32 & 95.62 & 96.80 & 57.81 & 92.57 & --- & --- & --- \\
  & SubspaceAD & \venue{CVPR'26} & \textbf{+DePrint-S1} & \textbf{94.51} & \textbf{97.39} & \textbf{95.65} & \textbf{96.86} & \textbf{58.08} & \textbf{92.69} & \pos{+0.17} & \pos{+0.06} & \pos{+0.12} \\
  & & & \cc +DePrint-S1+S2 & \cc 94.15 & \cc 97.17 & \cc 95.50 & \cc 95.88 & \cc 55.04 & \cc 91.28 & \cc\negv{$-$0.19} & \cc\negv{$-$0.92} & \cc\negv{$-$0.91} \\
\midrule
  & & & Baseline & 84.39 & 92.81 & 92.14 & 93.66 & 49.74 & 83.11 & --- & --- & --- \\
  & RD4AD & \venue{CVPR'22} & +DePrint-S1 & 87.72 & 94.33 & 93.01 & \textbf{95.24} & \textbf{52.91} & \textbf{86.54} & \pos{+3.33} & \pos{+1.58} & \pos{+2.32} \\
  \vcSix{Recon.} & & & \cc\textbf{+DePrint-S1+S2} & \cc\textbf{88.18} & \cc\textbf{94.63} & \cc\textbf{93.25} & \cc 95.21 & \cc 52.45 & \cc 85.61 & \cc\pos{+3.79} & \cc\pos{+1.55} & \cc\pos{+2.25} \\
\cmidrule(lr){2-13}
  & & & Baseline & 97.28 & 98.44 & 96.84 & 97.09 & 63.73 & \textbf{92.85} & --- & --- & --- \\
  & INP-Former & \venue{CVPR'25} & \textbf{+DePrint-S1} & 97.38 & 98.53 & \textbf{97.07} & 97.13 & \textbf{64.51} & 92.81 & \pos{+0.11} & +0.04 & \pos{+0.20} \\
  & & & \cc +DePrint-S1+S2 & \cc\textbf{97.47} & \cc\textbf{98.62} & \cc 96.97 & \cc\textbf{97.43} & \cc 64.04 & \cc\textbf{92.85} & \cc\pos{+0.19} & \cc\pos{+0.33} & \cc\pos{+0.19} \\
\midrule\midrule
\rowcolor{panelbg}\multicolumn{13}{l}{\textbf{VisA} \quad \textcolor{venuetext}{\textsf{\scriptsize 12 categories, 2/4-shot 6-run mean}}} \\
\midrule
  & & & Baseline & 87.78 & 88.14 & 85.23 & 97.04 & 39.90 & 85.56 & --- & --- & --- \\
  & AnomalyDINO & \venue{WACV'25} & +DePrint-S1 & 90.20 & 90.77 & \textbf{87.22} & 98.41 & \textbf{44.74} & 91.01 & \pos{+2.42} & \pos{+1.37} & \pos{+3.12} \\
  \vcSix{Memory} & & & \cc\textbf{+DePrint-S1+S2} & \cc\textbf{90.27} & \cc\textbf{90.94} & \cc 86.84 & \cc\textbf{98.51} & \cc 43.66 & \cc\textbf{92.15} & \cc\pos{+2.49} & \cc\pos{+1.47} & \cc\pos{+3.12} \\
\cmidrule(lr){2-13}
  & & & Baseline & 77.94 & 80.07 & 79.55 & 91.16 & 31.14 & 58.91 & --- & --- & --- \\
  & PatchCore & \venue{CVPR'22} & +DePrint-S1 & 80.19 & 81.91 & 80.80 & \textbf{93.78} & 32.26 & \textbf{67.31} & \pos{+2.25} & \pos{+2.62} & \pos{+2.91} \\
  & & & \cc\textbf{+DePrint-S1+S2} & \cc\textbf{80.39} & \cc\textbf{82.21} & \cc\textbf{81.12} & \cc 92.69 & \cc\textbf{32.59} & \cc 66.96 & \cc\pos{+2.45} & \cc\pos{+1.53} & \cc\pos{+2.87} \\
\midrule
  & & & Baseline & 67.12 & 71.80 & 76.86 & 94.03 & 18.13 & 70.64 & --- & --- & --- \\
  & PaDiM & \venue{ICPR'21} & \textbf{+DePrint-S1} & 74.56 & 78.28 & \textbf{78.24} & \textbf{96.49} & \textbf{28.53} & \textbf{78.63} & \pos{+7.44} & \pos{+2.46} & \pos{+6.02} \\
  & & & \cc +DePrint-S1+S2 & \cc\textbf{74.79} & \cc\textbf{78.72} & \cc 78.09 & \cc 94.73 & \cc 28.36 & \cc 68.52 & \cc\pos{+7.67} & \cc\pos{+0.70} & \cc\pos{+4.11} \\
\cmidrule(lr){2-13}
  & & & Baseline & 81.90 & 82.90 & 81.46 & 95.84 & 34.68 & 76.19 & --- & --- & --- \\
  Math. & FastRecon & \venue{ICCV'23} & +DePrint-S1 & 82.28 & 83.25 & 81.87 & \textbf{96.16} & 35.06 & 77.29 & \pos{+0.38} & \pos{+0.32} & \pos{+0.49} \\
  & & & \cc\textbf{+DePrint-S1+S2} & \cc\textbf{82.87} & \cc\textbf{84.00} & \cc\textbf{82.58} & \cc 96.05 & \cc\textbf{35.44} & \cc\textbf{77.39} & \cc\pos{+0.97} & \cc\pos{+0.21} & \cc\pos{+0.89} \\
\cmidrule(lr){2-13}
  & & & Baseline & 87.16 & 87.91 & 84.56 & 96.37 & 32.59 & 90.31 & --- & --- & --- \\
  & SubspaceAD & \venue{CVPR'26} & \textbf{+DePrint-S1} & 88.30 & 88.95 & \textbf{85.52} & 96.69 & \textbf{34.28} & \textbf{91.11} & \pos{+1.14} & \pos{+0.32} & \pos{+0.99} \\
  & & & \cc +DePrint-S1+S2 & \cc\textbf{88.74} & \cc\textbf{89.84} & \cc 85.46 & \cc\textbf{96.73} & \cc 31.52 & \cc 89.93 & \cc\pos{+1.58} & \cc\pos{+0.36} & \cc\pos{+0.55} \\
\midrule
  & & & Baseline & 73.04 & 76.49 & 78.63 & 94.07 & 24.08 & 70.40 & --- & --- & --- \\
  & RD4AD & \venue{CVPR'22} & \textbf{+DePrint-S1} & \textbf{79.26} & \textbf{81.98} & \textbf{80.71} & \textbf{96.45} & \textbf{32.58} & \textbf{79.01} & \pos{+6.22} & \pos{+2.38} & \pos{+5.55} \\
  \vcSix{Recon.} & & & \cc +DePrint-S1+S2 & \cc 79.07 & \cc\textbf{81.98} & \cc 80.68 & \cc 96.04 & \cc 32.57 & \cc 76.96 & \cc\pos{+6.03} & \cc\pos{+1.97} & \cc\pos{+5.10} \\
\cmidrule(lr){2-13}
  & & & Baseline & 95.46 & 95.42 & 91.87 & 97.51 & 48.92 & 93.47 & --- & --- & --- \\
  & INP-Former & \venue{CVPR'25} & +DePrint-S1 & 95.70 & 95.67 & 92.27 & 98.05 & \textbf{50.43} & 94.12 & \pos{+0.24} & \pos{+0.54} & \pos{+0.60} \\
  & & & \cc\textbf{+DePrint-S1+S2} & \cc\textbf{95.93} & \cc\textbf{95.99} & \cc\textbf{92.75} & \cc\textbf{98.30} & \cc 49.28 & \cc\textbf{94.56} & \cc\pos{+0.47} & \cc\pos{+0.80} & \cc\pos{+0.69} \\
\bottomrule
\end{tabular}}
\end{table*}

\subsection{Cross-Paradigm Few-Shot Benchmark}

\begin{table*}[t]
\centering
\renewcommand{\arraystretch}{1.13}
\setlength{\tabcolsep}{5.0pt}
\caption{\textbf{Zero-retraining per-category calibration on official full-shot multi-class SOTA checkpoints (\%).} Strictly aligned to official $R448\text{--}C392$ preprocessing and GT cropping (3 calibration seeds). \texttt{+DePrint-S1} uses exact streaming full-dataset Stage~1; \texttt{+DePrint-S1+S2} adds Stage~2 spatial Pareto. $|\Delta|\le 0.05$ shown in neutral black.}
\label{tab:main_fullshot}
\resizebox{\linewidth}{!}{%
\begin{tabular}{@{}ccclcccccccc@{}}
\toprule
 & & & & \multicolumn{3}{c}{\textbf{Image-Level}} & \multicolumn{3}{c}{\textbf{Pixel-Level}} & \multicolumn{2}{c}{\textbf{Gain}} \\
\cmidrule(lr){5-7}\cmidrule(lr){8-10}\cmidrule(lr){11-12}
\textbf{Dataset} & \textbf{Detector} & \textbf{Venue} & \textbf{Variant} & \textbf{I-AUC} & \textbf{I-AP} & \textbf{I-$F_1$} & \textbf{P-AUC} & \textbf{P-$F_1$} & \textbf{PRO} & \textbf{$\Delta$ I-AUC} & \textbf{$\Delta$ P-AUC} \\
\midrule
  & & & Baseline & \textbf{99.69} & 99.83 & 99.27 & 98.33 & 67.26 & \textbf{94.74} & --- & --- \\
  & Dinomaly & \venue{CVPR'25} & +DePrint-S1 & 99.68 & 99.82 & 99.18 & 98.29 & \textbf{67.69} & 94.60 & $-$0.02 & $-$0.04 \\
  \vcSix{MVTec-AD} & & & \cc +DePrint-S1+S2 & \cc\textbf{99.69} & \cc\textbf{99.86} & \cc\textbf{99.34} & \cc\textbf{98.37} & \cc 65.47 & \cc 94.16 & \cc +0.00 & \cc +0.04 \\
\cmidrule(lr){2-12}
  & & & Baseline & \textbf{99.67} & \textbf{99.88} & 99.20 & \textbf{98.44} & 67.63 & \textbf{94.80} & --- & --- \\
  & INP-Former & \venue{CVPR'25} & +DePrint-S1 & 99.65 & 99.86 & \textbf{99.26} & 98.41 & \textbf{67.86} & 94.68 & $-$0.02 & $-$0.03 \\
  & & & \cc +DePrint-S1+S2 & \cc 99.63 & \cc 99.87 & \cc 99.09 & \cc 98.42 & \cc 65.18 & \cc 93.88 & \cc $-$0.03 & \cc $-$0.03 \\
\midrule
  & & & Baseline & 98.85 & 99.00 & 96.51 & 98.75 & 51.80 & 95.28 & --- & --- \\
  & Dinomaly & \venue{CVPR'25} & \textbf{+DePrint-S1} & 98.77 & 98.89 & 96.34 & \textbf{98.94} & \textbf{52.45} & \textbf{95.51} & \negv{$-$0.08} & \pos{+0.19} \\
  \vcSix{VisA} & & & \cc +DePrint-S1+S2 & \cc\textbf{98.89} & \cc\textbf{99.08} & \cc\textbf{96.45} & \cc 98.93 & \cc 48.24 & \cc 93.33 & \cc +0.03 & \cc\pos{+0.17} \\
\cmidrule(lr){2-12}
  & & & Baseline & 98.90 & 99.02 & 96.57 & 98.96 & 51.18 & 95.08 & --- & --- \\
  & INP-Former & \venue{CVPR'25} & \textbf{+DePrint-S1} & 98.84 & 98.93 & 96.51 & \textbf{99.10} & \textbf{51.75} & \textbf{95.56} & \negv{$-$0.06} & \pos{+0.14} \\
  & & & \cc +DePrint-S1+S2 & \cc\textbf{99.11} & \cc\textbf{99.23} & \cc\textbf{96.87} & \cc 99.02 & \cc 47.76 & \cc 93.64 & \cc\pos{+0.21} & \cc\pos{+0.06} \\
\bottomrule
\end{tabular}}
\end{table*}

Table~\ref{tab:main_fewshot} presents our main few-shot benchmark across all seven detectors. \texttt{DePrint-S1} achieves strictly positive 6-metric mean gains on \textbf{14/14 detector-dataset pairs}, with the largest gains on VisA: PaDiM gains $+7.44$ I-AUC / $+6.02$ Avg., RD4AD gains $+6.22$ / $+5.55$, AnomalyDINO gains $+2.42$ / $+3.12$, and PatchCore gains $+2.25$ / $+2.91$. Even on the strongest official 2/4-shot reconstruction baseline (INP-Former, calibrated strictly on its exact 2/4 training images), \texttt{DePrint-S1} adds $+0.60$ Avg.\ ($+1.51$ P-$F_1$) on VisA and $+0.20$ Avg.\ on MVTec-AD, while \texttt{DePrint-S1+S2} raises VisA further to $+0.69$ Avg.\ ($+0.47$ I-AUC, $+0.80$ P-AUC, $+1.09$ PRO). Across all seven detectors on VisA, every single delta entry (42/42 cells) is positive. Qualitatively, Fig.~\ref{fig:overview}(c) illustrates this two-stage separation on VisA \texttt{capsules} (\texttt{bad/065}): Stage~1 suppresses widespread red false positives across normal capsules ($0.929\to 0.974$ pixel AUROC), and Stage~2 removes residual background clutter to isolate the defective capsule matching GT ($0.974\to 0.992$). At the same time, Table~\ref{tab:main_fewshot} transparently shows where Stage~2 trades pixel localization: on PaDiM VisA and SubspaceAD MVTec-AD, Stage~2 clips low-FPR pixel tails (reducing PRO or P-$F_1$), making \texttt{DePrint-S1} alone the stronger choice on those two models.

\subsection{Official Full-Shot Multi-Class SOTA}

\begin{figure*}[t]
\centering
\includegraphics[width=0.95\linewidth]{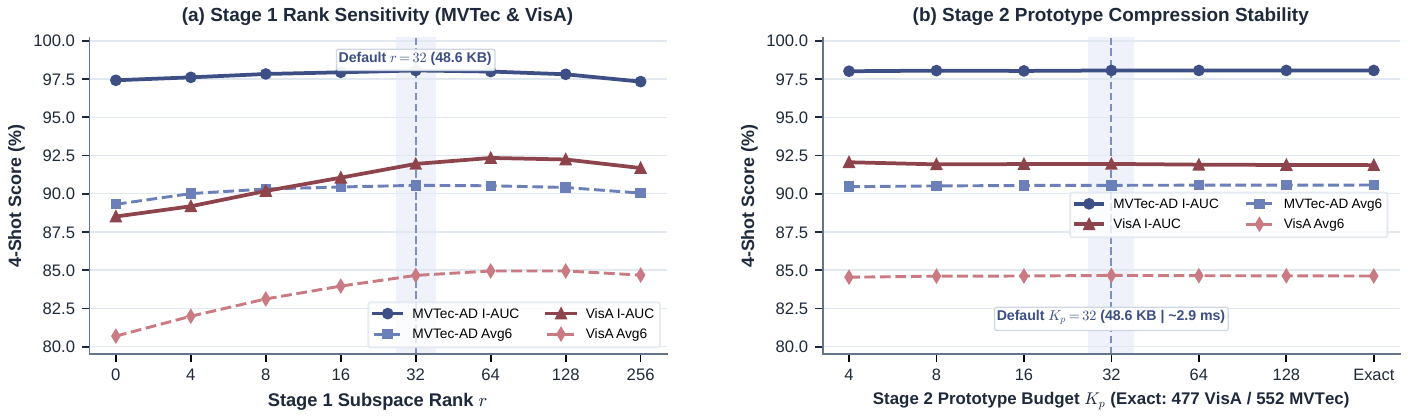}
\caption{\textbf{Hyperparameter stability across both MVTec-AD (15 classes) and VisA (12 classes) at 4-shot (AnomalyDINO, non-register \texttt{DINOv2-B/14}, $448\times 448$, seed 43).} (a) \textbf{Suite A (Rank $r$):} Default $r=32$ ($48.6$\,KB) achieves the exact peak on MVTec-AD ($98.06\%$ I-AUC / $90.55\%$ Avg6) and lies within $0.39$\,pp of peak $r=64$ on VisA at half the memory. (b) \textbf{Suite D (Prototype Budget $K_p$):} Performance is virtually flat from $K_p=4$ to exact uncompressed Pareto ($477$ VisA / $552$ MVTec prototypes), validating Proposition~\ref{prop:compress}.}
\label{fig:ablations}
\end{figure*}

Table~\ref{tab:main_fullshot} evaluates DePrint on saturated official full-shot multi-class checkpoints ($99.7\%$ MVTec-AD / $98.9\%$ VisA). As expected when full-dataset training already absorbs much of the support-induced coverage bias, overall performance is \textbf{near-neutral} on MVTec-AD ($|\Delta|\le 0.04$\,pp on I-AUC and P-AUC, with \texttt{DePrint-S1} slightly lifting P-$F_1$ by $+0.43$/$+0.23$\,pp). On VisA, the two stages cleanly specialize: \texttt{DePrint-S1} improves all three pixel metrics (Dinomaly P-$F_1$ $+0.65$, PRO $95.51\%$; INP-Former P-$F_1$ $+0.57$, PRO $95.56\%$), whereas \texttt{DePrint-S1+S2} pushes INP-Former VisA I-AUC from $98.90\%$ to $\mathbf{99.11\%}$ ($19\%$ relative error reduction) at the clear cost of lower P-$F_1$ ($-3.42$/$-3.56$\,pp) and PRO ($-1.44$/$-1.95$\,pp) due to hard zero-floor clipping on already-saturated pixel tails.

\subsection{Five-Suite Design Ablations}

To dissect every design choice in \texttt{DePrint-S1} and \texttt{DePrint-S1+S2}, we run five controlled ablation suites across all 27 categories of \textbf{MVTec-AD} (15 classes) and \textbf{VisA} (12 classes) at 4-shot (AnomalyDINO, non-register \texttt{DINOv2-B/14}, $448\times 448$, seed 43). Figure~\ref{fig:ablations} plots the rank and prototype sensitivity curves, and Appendix Table~\ref{tab:ablations} reports the complete breakdown for both datasets.

Key takeaways across all 27 categories (full numbers in Appendix Table~\ref{tab:ablations}):
\textbf{(A) Rank $r$:} Default $r=32$ ($48.6$\,KB) achieves the global peak on MVTec-AD ($98.06$ I-AUC / $90.55$ Avg6, declining to $97.33$ / $90.04$ at $r=256$) and stays within $0.39$\,pp I-AUC of the VisA peak $r=64$ ($91.95$ vs.\ $92.34$) at half the memory ($48.6$\,KB vs.\ $96.6$\,KB).
\textbf{(B) Centering:} Uncentered SVD and centered PCA are empirically indistinguishable on both datasets (VisA $84.66$ vs.\ $84.65$; MVTec $90.55$ vs.\ $90.54$ Avg6, difference $<0.02$\,pp), whereas mean-only subtraction fails ($81.61$ VisA / $89.80$ MVTec) and partial projection ($\alpha=0.5$) underperforms; uncentered SVD offers an origin-anchored formulation with zero extra mean parameters.
\textbf{(C) Spatial kernel:} Removing spatial context ($1\times 1$) collapses Avg6 on both datasets ($73.96$ VisA, $-10.70$\,pp; $86.39$ MVTec, $-4.16$\,pp, with MVTec PRO dropping from $93.20$ to $82.86$); all $3\times 3$ weighting schemes lie within $0.14$\,pp on VisA ($84.57$--$84.71$) and $0.09$\,pp on MVTec ($90.50$--$90.59$), while $5\times 5$ (25-D) slightly degrades both ($84.26$ / $90.34$).
\textbf{(D) Prototype budget $K_p$:} Performance is flat from $K_p=4$ to exact uncompressed Pareto on both VisA ($477.1$ prototypes, $84.54$--$84.66$) and MVTec ($551.6$ prototypes, $90.46$--$90.57$); $K_p=32$ prunes $93.3\%$--$94.2\%$ of spatial prototypes ($48.6$\,KB total, $2.69$\,ms VisA / $3.03$\,ms MVTec) while matching exact Pareto within $0.02$--$0.04$\,pp.
\textbf{(E) Geometry and causal order:} Applying Stage~2 without Stage~1 drops below baseline on both datasets (VisA $80.69$ vs.\ $81.62$; MVTec $89.31$ vs.\ $89.74$ Avg6), confirming our causal ordering. Replacing one-sided Pareto with two-sided $\ell_2$ ($64.28$ VisA / $79.34$ MVTec) or convex Mahalanobis ($81.94$ VisA / $88.24$ MVTec) degrades sharply, confirming Corollary~\ref{cor:nonconvex}. Finally, Suite~E exposes the mechanism behind Stage~2's P-$F_1$ trade-off: Stage~1 drives the bulk of the gain ($+2.90$ VisA / $+0.78$ MVTec Avg6), while Stage~2's hard zero floor $\max(x - F_{c,k}, \mathbf{0})$ zeros out every block dominated by a frontier prototype. This removes sparse normal clutter (lifting I-AUC by $+0.44$ / $+0.53$\,pp and VisA PRO by $+1.11$\,pp), but also clips weak defect-boundary pixels below $F_{c,k}$, reducing recall at the P-$F_1$ threshold ($-1.86$ VisA / $-1.46$ MVTec; $-3.42$/$-3.56$\,pp on full-shot VisA). Practically, we recommend \texttt{DePrint-S1} alone when pixel $F_1$-max is prioritized or on whitened/saturated detectors (PaDiM, SubspaceAD, full-shot localization), and \texttt{DePrint-S1+S2} for few-shot image detection and AU-PRO. \textbf{(F) 1-Shot Augmented-View Calibration (Appendix~\ref{app:robustness}, Table~\ref{tab:suite_f}):} At $K=1$, mild augmented-view hold-out ($6$ deterministic views) recovers over half of the 4-shot LOIO drift subspace (Grassmann overlap $0.545$ VisA / $0.522$ MVTec vs.\ $0.042$ null), boosting 1-shot Avg6 by $+4.02$\,pp on VisA (\texttt{DePrint-S1}, $76.07\to 80.09$) and $+1.65$\,pp on MVTec (\texttt{DePrint-S1+S2}, $86.90\to 88.55$) without overfitting at $r=32$. \textbf{Pure-Text Zero-Shot VLM (Appendix~\ref{app:vlm}, Table~\ref{tab:vlm_anomalyclip}):} On official zero-shot AnomalyCLIP (\texttt{ViT-L/14@336px}), $4$-shot post-hoc calibration of cross-modal residuals $q - t_{\mathrm{norm}}$ reveals an ultra-low-rank modality-gap subspace ($r_{\mathrm{eff}}/D = 0.004$, $E_{32} \approx 95.6\%$--$95.9\%$ for both normal and anomaly); projecting out $U_c$ (\texttt{S1-Res}) exposes directional defect discriminability in $U_c^\perp$ and lifts Avg6 from $81.50\to 88.41$ on MVTec-AD and $75.08\to 78.74$ on VisA, whereas naively orthogonalizing the learned prompt axis (\texttt{S1-Axis}) degrades VisA ($75.08\to 70.47$) and Stage~2 alone is near-neutral ($81.69$ / $74.77$).

% =========================================================================
\section{Limitations and Conclusion}

\noindent\textbf{Limitations.} Four practical boundaries should be noted.
First, DePrint delivers its largest gains in few-shot regimes where finite-support coverage bias is strongest, whereas on saturated full-shot multi-class checkpoints ($>99\%$ MVTec-AD / $\approx 99\%$ VisA) its net effect is near-neutral.
Second, Stage~2's hard zero floor $\max(x - F_{c,k}, \mathbf{0})$ introduces a measurable trade-off: while it improves few-shot image detection and AU-PRO on most detectors, it can lower pixel $F_1$-max (e.g., $45.50\to 43.64$ on VisA 4-shot AnomalyDINO) and clip low-FPR pixel tails on saturated full-shot models, PaDiM, and SubspaceAD; deploying \texttt{DePrint-S1} alone avoids this trade-off when pixel $F_1$-max is prioritized.
Third, at $K=1$, cross-image LOIO is undefined and augmented-view hold-out captures encoder bias (Grassmann overlap $0.522$--$0.545$, Appendix~\ref{app:robustness}) but misses cross-instance variance (e.g., Stage~2 lowers 1-shot VisA I-AUC despite lifting pixel metrics). Moreover, severe unaugmented pose shifts exceed Eq.~\ref{eq:taylor}'s local tangent approximation and remain bounded by host detector invariance.
Fourth, our five-suite design ablation covers all 27 MVTec-AD and VisA categories at 4-shot (seed 43); multi-seed 8-shot scaling, rotation-augmented calibration, and broader multi-view benchmarks (such as BTAD~\cite{mishra2021btad} and Real-IAD~\cite{wang2024realiad}) are left to supplementary and future evaluation.

\noindent\textbf{Conclusion.} Every visual anomaly detector paired with a finite normal support set leaves an intrinsic residual fingerprint composed of a low-rank channel drift subspace and a non-convex spatial Pareto order ideal. Canceling both in causal order via \textbf{DePrint} requires no training, $48.6$\,KB per class, and $2.69$\,ms per image, delivering consistent few-shot gains across seven detectors spanning memory-bank, mathematical modeling, and reconstruction paradigms while offering a transparent Stage-1/Stage-2 operating choice on saturated full-shot models.

% =========================================================================

\clearpage
\onecolumn
\appendix

\section{Complete 27-Class Five-Suite Design Ablation (43 Configurations)}
\label{app:ablations}

Table~\ref{tab:ablations} reports all 43 configurations across the five design and hyperparameter ablation suites on \textbf{MVTec-AD} (15 classes) and \textbf{VisA} (12 classes) at 4-shot (AnomalyDINO, non-register \texttt{DINOv2-B/14}, $448\times 448$, seed 43), including float16 storage footprint (KB), mean spatial prototype count ($K_p$), and measured GPU latency (ms/img) on Suite~D. Every number is reported verbatim without omission.

\begin{table}[!htbp]
\centering
\renewcommand{\arraystretch}{1.01}
\setlength{\tabcolsep}{3.0pt}
\caption{\textbf{Complete 43-configuration five-suite ablation on MVTec-AD (15 classes) and VisA (12 classes) at 4-shot (AnomalyDINO, non-register \texttt{DINOv2-B/14}, $448\times 448$, seed 43, \%).} Default DePrint configuration ($r=32$, uncentered SVD, Gaussian $3\times 3$, $K_p=32$, one-sided Pareto) is shaded. Cost columns report float16 KB and MVTec / VisA prototype count $K_p$ and latency (ms/img).}
\label{tab:ablations}
\resizebox{\linewidth}{!}{%
\begin{tabular}{@{}lccccccc@{\hspace{7pt}}ccccccc@{\hspace{6pt}}ccc@{}}
\toprule
 & \multicolumn{7}{c}{\textbf{MVTec-AD (15 classes, 4-shot, seed 43)}} & \multicolumn{7}{c}{\textbf{VisA (12 classes, 4-shot, seed 43)}} & \multicolumn{3}{c}{\textbf{Cost (MVTec / VisA)}} \\
\cmidrule(lr){2-8}\cmidrule(lr){9-15}\cmidrule(lr){16-18}
\textbf{Configuration} & \textbf{I-AUC} & \textbf{I-AP} & \textbf{I-$F_1$} & \textbf{P-AUC} & \textbf{P-$F_1$} & \textbf{PRO} & \textbf{Avg6} & \textbf{I-AUC} & \textbf{I-AP} & \textbf{I-$F_1$} & \textbf{P-AUC} & \textbf{P-$F_1$} & \textbf{PRO} & \textbf{Avg6} & \textbf{KB} & \textbf{$K_p$} & \textbf{ms} \\
\midrule
\rowcolor{panelbg}\multicolumn{18}{l}{\textbf{Suite A: Stage-1 Subspace Rank $r \in \{0,4,8,16,32,64,128,256\}$ (S1-only vs.\ Full S1+S2)}} \\
Rank $r=0$ (S1) & 97.02 & 98.20 & 96.49 & 96.31 & 58.47 & 91.97 & 89.74 & 89.51 & 89.80 & 86.25 & 97.31 & 40.61 & 86.21 & 81.62 & 0.0 / 0.0 & --- & --- \\
Rank $r=0$ (S1+S2) & 97.42 & 98.58 & 96.87 & 96.32 & 56.28 & 90.39 & 89.31 & 88.51 & 89.36 & 85.14 & 96.74 & 39.21 & 85.17 & 80.69 & 0.5 / 0.6 & 30.7 / 32.0 & --- \\
Rank $r=4$ (S1) & 97.16 & 98.29 & 96.51 & 96.59 & 59.00 & 92.42 & 90.00 & 89.72 & 90.19 & 86.70 & 97.75 & 41.94 & 87.91 & 82.37 & 6.0 / 6.0 & --- & --- \\
Rank $r=4$ (S1+S2) & 97.61 & 98.68 & 97.06 & 97.02 & 57.30 & 92.36 & 90.01 & 89.18 & 90.06 & 85.56 & 97.34 & 40.46 & 89.34 & 81.99 & 6.6 / 6.6 & 32.0 / 32.0 & --- \\
Rank $r=8$ (S1) & 97.28 & 98.41 & 96.75 & 96.72 & 59.31 & 92.68 & 90.19 & 90.27 & 90.96 & 86.96 & 98.07 & 43.14 & 89.27 & 83.11 & 12.0 / 12.0 & --- & --- \\
Rank $r=8$ (S1+S2) & 97.83 & 98.84 & 97.39 & 97.10 & 57.82 & 92.87 & 90.31 & 90.18 & 91.00 & 86.49 & 97.97 & 42.01 & 91.08 & 83.12 & 12.6 / 12.6 & 32.0 / 32.0 & --- \\
Rank $r=16$ (S1) & 97.41 & 98.45 & 96.78 & 96.89 & 59.74 & 92.98 & 90.38 & 90.97 & 91.53 & 87.30 & 98.35 & 44.51 & 90.50 & 83.86 & 24.0 / 24.0 & --- & --- \\
Rank $r=16$ (S1+S2) & 97.94 & 98.89 & 97.20 & 97.23 & 58.28 & 93.10 & 90.44 & 91.05 & 91.67 & 87.33 & 98.42 & 43.35 & 91.93 & 83.96 & 24.6 / 24.6 & 32.0 / 32.0 & --- \\
Rank $r=32$ (S1) & 97.53 & 98.61 & 96.84 & 97.04 & \textbf{59.94} & 93.19 & 90.52 & 91.51 & 92.01 & 87.88 & 98.59 & 45.50 & 91.63 & 84.52 & 48.0 / 48.0 & --- & --- \\
\rowcolor{deprinttint}Rank $r=32$ (S1+S2, \textbf{Ours}) & \textbf{98.06} & \textbf{98.98} & 97.27 & 97.30 & 58.48 & \textbf{93.20} & \textbf{90.55} & 91.95 & 92.48 & 88.49 & 98.66 & 43.64 & 92.74 & 84.66 & 48.6 / 48.6 & 32.0 / 32.0 & --- \\
Rank $r=64$ (S1) & 97.37 & 98.53 & 96.78 & 97.09 & 59.66 & 93.11 & 90.42 & 91.89 & 92.25 & 88.42 & 98.71 & \textbf{45.67} & 92.28 & 84.87 & 96.0 / 96.0 & --- & --- \\
Rank $r=64$ (S1+S2) & 97.99 & 98.97 & 97.24 & \textbf{97.33} & 58.49 & 93.11 & 90.52 & \textbf{92.34} & \textbf{92.79} & \textbf{88.61} & 98.77 & 44.03 & 93.16 & \textbf{84.95} & 96.6 / 96.6 & 32.0 / 32.0 & --- \\
Rank $r=128$ (S1) & 97.17 & 98.42 & 96.61 & 97.02 & 59.41 & 92.87 & 90.25 & 91.64 & 91.91 & 87.93 & 98.76 & 45.63 & 92.45 & 84.72 & 192.0 / 192.0 & --- & --- \\
Rank $r=128$ (S1+S2) & 97.81 & 98.84 & 97.22 & 97.25 & 58.47 & 92.85 & 90.41 & 92.24 & 92.58 & 88.41 & \textbf{98.82} & 44.36 & \textbf{93.30} & \textbf{84.95} & 192.6 / 192.6 & 32.0 / 32.0 & --- \\
Rank $r=256$ (S1) & 96.72 & 98.08 & 96.35 & 96.93 & 58.75 & 92.48 & 89.89 & 90.99 & 91.12 & 87.55 & 98.75 & 45.38 & 92.50 & 84.38 & 384.0 / 384.0 & --- & --- \\
Rank $r=256$ (S1+S2) & 97.33 & 98.57 & 97.12 & 97.09 & 57.82 & 92.34 & 90.04 & 91.68 & 92.03 & 88.10 & 98.79 & 44.31 & 93.17 & 84.68 & 384.6 / 384.6 & 32.0 / 32.0 & --- \\
\midrule
\rowcolor{panelbg}\multicolumn{18}{l}{\textbf{Suite B: Stage-1 Centering and Projection Strength (Prop.~\ref{prop:svd})}} \\
Uncentered SVD ($r=32$, S1) & 97.53 & 98.61 & 96.84 & 97.04 & \textbf{59.94} & 93.19 & 90.52 & 91.51 & 92.01 & 87.88 & 98.59 & \textbf{45.50} & 91.63 & 84.52 & 48.0 / 48.0 & --- & --- \\
\rowcolor{deprinttint}Uncentered SVD ($r=32$, S1+S2) & \textbf{98.06} & \textbf{98.98} & \textbf{97.27} & \textbf{97.30} & 58.48 & \textbf{93.20} & \textbf{90.55} & \textbf{91.95} & \textbf{92.48} & \textbf{88.49} & \textbf{98.66} & 43.64 & \textbf{92.74} & \textbf{84.66} & 48.6 / 48.6 & 32.0 / 32.0 & --- \\
Centered PCA ($r=32$, S1) & 97.52 & 98.59 & 96.87 & 97.04 & \textbf{59.94} & 93.19 & 90.53 & 91.51 & 92.01 & 87.86 & 98.59 & \textbf{45.50} & 91.64 & 84.52 & 48.0 / 48.0 & --- & --- \\
Centered PCA ($r=32$, S1+S2) & 98.04 & 98.97 & \textbf{97.27} & \textbf{97.30} & 58.47 & 93.19 & 90.54 & \textbf{91.95} & \textbf{92.48} & 88.45 & \textbf{98.66} & 43.63 & 92.70 & 84.65 & 48.6 / 48.6 & 32.0 / 32.0 & --- \\
Mean-Only ($r=0$, S1) & 97.05 & 98.21 & 96.61 & 96.34 & 58.57 & 92.01 & 89.80 & 89.48 & 89.77 & 86.21 & 97.31 & 40.66 & 86.23 & 81.61 & 0.0 / 0.0 & --- & --- \\
Mean-Only ($r=0$, S1+S2) & 97.46 & 98.60 & 96.93 & 96.40 & 56.39 & 90.61 & 89.40 & 88.50 & 89.35 & 85.29 & 96.75 & 39.35 & 85.33 & 80.76 & 0.5 / 0.6 & 30.7 / 32.0 & --- \\
Partial Proj.\ ($\alpha=0.5$, S1) & 97.47 & 98.51 & 96.77 & 96.90 & 59.74 & 92.98 & 90.39 & 91.07 & 91.51 & 87.42 & 98.34 & 44.33 & 90.41 & 83.85 & 48.0 / 48.0 & --- & --- \\
Partial Proj.\ ($\alpha=0.5$, S1+S2) & 97.94 & 98.89 & 97.18 & 97.23 & 57.96 & 93.05 & 90.38 & 91.23 & 91.80 & 87.55 & 98.42 & 42.31 & 91.77 & 83.85 & 48.6 / 48.6 & 32.0 / 32.0 & --- \\
\midrule
\rowcolor{panelbg}\multicolumn{18}{l}{\textbf{Suite C: Stage-2 Spatial Neighborhood Kernel and Window Size}} \\
Point $1\times 1$ (no window) & 96.72 & 98.25 & 95.99 & 92.48 & 52.04 & 82.86 & 86.39 & 86.93 & 88.28 & 83.65 & 87.44 & 35.58 & 61.87 & 73.96 & 48.0 / 48.0 & 1.0 / 1.0 & --- \\
\rowcolor{deprinttint}Gaussian $3\times 3$ (\textbf{Ours}) & 98.06 & 98.98 & 97.27 & 97.30 & 58.48 & 93.20 & 90.55 & 91.95 & 92.48 & 88.49 & 98.66 & 43.64 & 92.74 & 84.66 & 48.6 / 48.6 & 32.0 / 32.0 & --- \\
Uniform $3\times 3$ & \textbf{98.10} & \textbf{99.04} & 97.26 & 97.30 & 58.30 & 93.02 & 90.50 & 91.92 & 92.48 & 88.29 & 98.66 & 43.51 & 92.53 & 84.57 & 48.6 / 48.6 & 32.0 / 32.0 & --- \\
Cross $3\times 3$ (5-pt) & 98.07 & 98.97 & \textbf{97.40} & 97.14 & \textbf{58.58} & \textbf{93.38} & \textbf{90.59} & 91.88 & 92.32 & 88.37 & 98.53 & \textbf{44.03} & 92.75 & 84.65 & 48.3 / 48.3 & 32.0 / 31.6 & --- \\
Sharp $3\times 3$ (center 8) & 98.06 & 98.97 & 97.23 & 97.30 & 58.46 & 93.25 & 90.55 & \textbf{92.06} & \textbf{92.55} & \textbf{88.50} & \textbf{98.67} & 43.69 & \textbf{92.82} & \textbf{84.71} & 48.6 / 48.6 & 32.0 / 32.0 & --- \\
Gaussian $5\times 5$ (25-D) & 98.03 & 98.97 & 97.31 & \textbf{97.38} & 57.69 & 92.69 & 90.34 & 91.56 & 92.15 & 88.04 & 98.64 & 42.87 & 92.30 & 84.26 & 49.6 / 49.6 & 32.0 / 32.0 & --- \\
\midrule
\rowcolor{panelbg}\multicolumn{18}{l}{\textbf{Suite D: Stage-2 Spatial Prototype Budget $K_p$ and Measured GPU Latency (Prop.~\ref{prop:compress})}} \\
$K_p = 4$ & 98.01 & 98.95 & 97.23 & 97.24 & 58.18 & 93.14 & 90.46 & \textbf{92.06} & \textbf{92.60} & 88.32 & 98.58 & 42.94 & 92.76 & 84.54 & 48.1 / 48.1 & 4.0 / 4.0 & 2.93 / 2.56 \\
$K_p = 8$ & 98.05 & 98.97 & 97.19 & 97.28 & 58.40 & 93.19 & 90.51 & 91.92 & 92.52 & 88.44 & 98.63 & 43.31 & \textbf{92.81} & 84.61 & 48.1 / 48.1 & 8.0 / 8.0 & 2.86 / 2.64 \\
$K_p = 16$ & 98.03 & 98.96 & 97.28 & 97.29 & 58.49 & \textbf{93.21} & 90.54 & 91.94 & 92.48 & 88.37 & 98.65 & 43.56 & 92.74 & 84.62 & 48.3 / 48.3 & 16.0 / 16.0 & 2.87 / 2.57 \\
\rowcolor{deprinttint}$K_p = 32$ (\textbf{Ours}) & \textbf{98.06} & 98.98 & 97.27 & 97.30 & 58.48 & 93.20 & 90.55 & 91.95 & 92.48 & \textbf{88.49} & 98.66 & 43.64 & 92.74 & \textbf{84.66} & 48.6 / 48.6 & 32.0 / 32.0 & 3.03 / 2.69 \\
$K_p = 64$ & \textbf{98.06} & \textbf{98.99} & \textbf{97.29} & 97.31 & 58.52 & 93.19 & 90.56 & 91.90 & 92.43 & 88.48 & 98.66 & 43.71 & 92.68 & 84.64 & 49.1 / 49.1 & 64.0 / 64.0 & 2.92 / 3.30 \\
$K_p = 128$ & \textbf{98.06} & \textbf{98.99} & \textbf{97.29} & 97.31 & 58.53 & 93.19 & 90.56 & 91.88 & 92.40 & 88.41 & \textbf{98.67} & 43.79 & 92.65 & 84.63 & 50.2 / 50.2 & 128.0 / 128.0 & 2.89 / 3.40 \\
Exact Pareto (uncompressed) & \textbf{98.06} & 98.98 & \textbf{97.29} & \textbf{97.32} & \textbf{58.55} & 93.20 & \textbf{90.57} & 91.87 & 92.39 & 88.34 & \textbf{98.67} & \textbf{43.81} & 92.65 & 84.62 & 57.7 / 56.4 & 551.6 / 477.1 & 2.81 / 2.68 \\
\midrule
\rowcolor{panelbg}\multicolumn{18}{l}{\textbf{Suite E: Stage-2 Spatial Geometry and Causal Stage Ordering (Prop.~\ref{prop:pareto}, Cor.~\ref{cor:nonconvex})}} \\
Baseline ($r=0$) & 97.02 & 98.20 & 96.49 & 96.31 & 58.47 & 91.97 & 89.74 & 89.51 & 89.80 & 86.25 & 97.31 & 40.61 & 86.21 & 81.62 & 0.0 / 0.0 & --- & --- \\
Stage~1 Only (\texttt{DePrint-S1}) & 97.53 & 98.61 & 96.84 & 97.04 & \textbf{59.94} & 93.19 & 90.52 & 91.51 & 92.01 & 87.88 & 98.59 & \textbf{45.50} & 91.63 & 84.52 & 48.0 / 48.0 & --- & --- \\
Stage~2 Only (no S1) & 97.42 & 98.58 & 96.87 & 96.32 & 56.28 & 90.39 & 89.31 & 88.51 & 89.36 & 85.14 & 96.74 & 39.21 & 85.17 & 80.69 & 0.5 / 0.6 & 30.7 / 32.0 & --- \\
\rowcolor{deprinttint}S1 $\to$ S2 Pareto (\textbf{Ours}) & \textbf{98.06} & \textbf{98.98} & \textbf{97.27} & \textbf{97.30} & 58.48 & \textbf{93.20} & \textbf{90.55} & \textbf{91.95} & \textbf{92.48} & \textbf{88.49} & \textbf{98.66} & 43.64 & \textbf{92.74} & \textbf{84.66} & 48.6 / 48.6 & 32.0 / 32.0 & --- \\
S1 $\to$ S2 Two-Sided $\ell_2$ & 85.64 & 92.59 & 90.95 & 85.13 & 47.01 & 74.72 & 79.34 & 69.77 & 77.36 & 77.19 & 77.51 & 30.83 & 53.04 & 64.28 & 48.6 / 48.6 & 32.0 / 32.0 & --- \\
S1 $\to$ S2 Convex Maha-9 & 95.54 & 97.48 & 95.63 & 95.91 & 53.58 & 91.30 & 88.24 & 89.31 & 90.30 & 86.25 & 97.23 & 37.71 & 90.84 & 81.94 & 48.0 / 48.0 & 0.0 / 0.0 & --- \\
\bottomrule
\end{tabular}}
\end{table}

\clearpage
\section{1-Shot Applicability and Overfitting Check (Suite F)}
\label{app:robustness}

This experiment runs on AnomalyDINO with non-register \texttt{DINOv2-B/14} ($448\times 448$, seed 43), matching the exact protocol of Appendix~\ref{app:ablations}.

\subsection*{Suite F: 1-Shot Augmented-View Hold-Out and Overfitting Check}
At $K=1$, cross-image LOIO is undefined because holding out the sole normal image leaves an empty support set. We instead generate six deterministic mild perturbations of the single support image ($\pm 5^\circ$ border-filled rotation, $\pm 4$\,px shift, and $\pm 8\%$ brightness/contrast) and score them against the unaugmented 1-shot memory bank to fit $U_c$ and $F_c$. Table~\ref{tab:suite_f} reports the full rank sweep ($r \in \{0,8,16,32,64\}$) on all 12 VisA and 15 MVTec-AD categories, together with the mean Grassmann overlap $\frac{1}{r}\|U_{1\text{-shot}}^\top U_{4\text{-shot}}\|_F^2$ against standard 4-shot cross-image LOIO ($r=32$; isotropic null $= 32/768 = 0.042$).

Three observations follow directly from Table~\ref{tab:suite_f}:
(i) \textbf{Strong subspace recovery:} 1-shot augmented-view hold-out achieves a mean Grassmann overlap of $\mathbf{0.545}$ on VisA and $\mathbf{0.522}$ on MVTec-AD ($12\times$--$13\times$ above the $0.042$ null), showing that $U_c$ captures genuine encoder/detector drift rather than memorizing single-image background.
(ii) \textbf{No overfitting at default $r=32$:} On VisA, \texttt{DePrint-S1} improves monotonically from $76.07$ ($r=0$) to $80.09$ Avg6 at $r=32$ ($+4.02$\,pp; I-AUC $+2.84$, P-$F_1$ $+6.65$, PRO $+8.06$) and $80.40$ at $r=64$; on MVTec-AD, both \texttt{DePrint-S1} ($88.28$, $+1.38$) and \texttt{DePrint-S1+S2} ($88.55$, $+1.65$) peak cleanly at $r=32$ before dipping slightly at $r=64$ ($87.97$ / $88.14$).
(iii) \textbf{1-shot Stage-2 nuance:} On MVTec-AD, Stage~2 improves all six metrics over Stage~1; on VisA, Stage~2 lifts pixel metrics (P-AUC $97.60$, P-$F_1$ $41.84$, PRO $88.45$) while lowering I-AUC ($84.33\to 82.34$) because synthetic views of a single image under-cover cross-instance spatial variance.

\begin{table}[!htbp]
\centering
\renewcommand{\arraystretch}{1.08}
\setlength{\tabcolsep}{4.8pt}
\caption{\textbf{Suite F: 1-shot augmented-view calibration and rank sweep on VisA (12 classes) and MVTec-AD (15 classes) (AnomalyDINO, non-register \texttt{DINOv2-B/14}, $448\times 448$, seed 43, \%).} Mean Grassmann overlap between $U_{1\text{-shot}}(r=32)$ and 4-shot LOIO $U_{4\text{-shot}}(r=32)$ is \textbf{0.545} (VisA) and \textbf{0.522} (MVTec-AD) versus $0.042$ isotropic null.}
\label{tab:suite_f}
\resizebox{\linewidth}{!}{%
\begin{tabular}{@{}lccccccc@{\hspace{10pt}}ccccccc@{}}
\toprule
 & \multicolumn{7}{c}{\textbf{VisA (12 classes, 1-shot, Grassmann = 0.545)}} & \multicolumn{7}{c}{\textbf{MVTec-AD (15 classes, 1-shot, Grassmann = 0.522)}} \\
\cmidrule(lr){2-8}\cmidrule(lr){9-15}
\textbf{Config} & \textbf{I-AUC} & \textbf{I-AP} & \textbf{I-$F_1$} & \textbf{P-AUC} & \textbf{P-$F_1$} & \textbf{PRO} & \textbf{Avg6} & \textbf{I-AUC} & \textbf{I-AP} & \textbf{I-$F_1$} & \textbf{P-AUC} & \textbf{P-$F_1$} & \textbf{PRO} & \textbf{Avg6} \\
\midrule
$r=0$ (Baseline) & 81.49 & 83.26 & 82.53 & 95.16 & 34.67 & 79.32 & 76.07 & 93.25 & 96.16 & 94.34 & 94.64 & 54.20 & 88.81 & 86.90 \\
$r=8$ (S1) & 82.96 & 84.71 & 83.00 & 96.26 & 37.64 & 83.08 & 77.94 & 93.38 & 96.33 & 94.58 & 95.47 & 55.86 & 90.07 & 87.62 \\
$r=8$ (S1+S2) & 80.84 & 83.05 & 82.43 & 96.44 & 38.25 & 85.68 & 77.78 & 93.56 & 96.67 & 95.39 & 96.14 & 56.13 & 90.23 & 88.02 \\
$r=16$ (S1) & 84.43 & 86.31 & 83.60 & 97.00 & 39.85 & 85.54 & 79.46 & 93.67 & 96.55 & 94.56 & 95.89 & 56.42 & 90.71 & 87.97 \\
$r=16$ (S1+S2) & 81.93 & 84.19 & 82.49 & 97.11 & 40.39 & 87.30 & 78.90 & 93.82 & 96.77 & 95.40 & 96.47 & 56.65 & 90.81 & 88.32 \\
\rowcolor{deprinttint}$r=32$ (Default S1) & 84.33 & 86.28 & 83.75 & 97.51 & 41.32 & 87.38 & 80.09 & 94.00 & 96.67 & 94.62 & 96.33 & 56.77 & 91.28 & 88.28 \\
\rowcolor{deprinttint}$r=32$ (Default S1+S2) & 82.34 & 84.47 & 83.00 & 97.60 & \textbf{41.84} & 88.45 & 79.62 & \textbf{94.04} & \textbf{96.86} & \textbf{95.44} & 96.81 & \textbf{56.86} & 91.29 & \textbf{88.55} \\
$r=64$ (S1) & \textbf{84.59} & \textbf{86.34} & \textbf{83.95} & 97.81 & 41.53 & 88.21 & \textbf{80.40} & 93.12 & 96.11 & 94.61 & 96.50 & 56.14 & \textbf{91.34} & 87.97 \\
$r=64$ (S1+S2) & 82.38 & 84.38 & 83.30 & \textbf{97.89} & 41.74 & \textbf{88.98} & 79.78 & 93.32 & 96.24 & 95.22 & \textbf{96.83} & 56.01 & 91.19 & 88.14 \\
\bottomrule
\end{tabular}}
\end{table}

\clearpage
\section{Pure-Text Zero-Shot VLM Extension: Official AnomalyCLIP}
\label{app:vlm}

To test whether intrinsic residual fingerprints extend to pure text-prompted vision-language detectors, we evaluate official \textbf{AnomalyCLIP}~\cite{zhou2024anomalyclip} (\texttt{ViT-L/14@336px}, released \texttt{epoch15} cross-dataset checkpoints: VisA-trained evaluated on MVTec-AD, MVTec-trained evaluated on VisA). Inference remains strictly zero-shot with respect to the host detector (no visual memory bank and zero weight updates); we use only $K=4$ normal target images (seed 43) solely to fit post-hoc DePrint calibration $(U_c, F_c)$. All six configurations are evaluated under our unified 6-metric suite (\texttt{Base (Official)} denotes our 6-metric reproduction of the official checkpoint rather than numbers copied from the original paper).

\noindent\textbf{Diagnostic findings (different channel regime from DINOv2).}
On cross-modal patch residuals $\Delta_{\mathrm{txt}}^{(l)} = q^{(l)} - t_{\mathrm{norm}}$, AnomalyCLIP exhibits extreme low-rank concentration: $r_{\mathrm{eff}}/D = \mathbf{0.004}$ on both MVTec-AD and VisA, with $E_{32}$ (normal / anomaly) $= \mathbf{95.9\% / 95.4\%}$ on MVTec-AD and $\mathbf{95.6\% / 95.3\%}$ on VisA. Unlike DINOv2/INP-Former where normal energy substantially exceeds anomaly energy inside $U_c$, here both normal and anomalous patches share $\approx 95\%$ energy inside $U_c$ because of the dominant CLIP vision-text modality gap and category-wide feature cone. Yet removing $U_c$ via \texttt{S1-Res} still dramatically sharpens separation (MVTec Avg6 $81.50 \to \mathbf{88.41}$, P-AUC $+7.57$, P-$F_1$ $+20.00$; VisA Avg6 $75.08 \to \mathbf{78.74}$, P-$F_1$ $+9.29$): stripping the shared $95\%$ modality-gap norm reveals that true defect discriminability concentrates directionally inside the remaining $\sim 4$--$5\%$ orthogonal complement $U_c^\perp$. In the spatial 9-D diagnostic on raw text score maps, $r_{\mathrm{eff},9} = 0.368$ (MVTec) / $0.272$ (VisA) and one-sided Pareto outperforms both 1-D scalars and convex Mahalanobis on TPR@5\%FPR ($73.6\%$ vs.\ $70.8\% / 71.7\%$ on MVTec; $53.1\%$ vs.\ $44.9\% / 48.2\%$ on VisA), though overall text-map separation is weaker than DINOv2 detectors.

\noindent\textbf{Negative and neutral regimes (transparent boundary).}
Table~\ref{tab:vlm_anomalyclip} highlights two critical failure/neutral modes that must be distinguished from positive gains:
(i) \textbf{Prompt-axis orthogonalization (\texttt{S1-Axis}) fails on VisA:} while orthogonalizing the learned prompt axis $(t_{\mathrm{anom}} - t_{\mathrm{norm}})$ against $U_c$ raises MVTec I-AUC ($90.52\to 93.61$), it severely degrades pixel localization on both datasets and drops VisA Avg6 from $75.08$ to $70.47$ (P-AUC $94.41\to 83.96$, PRO $84.25\to 67.95$), showing that the learned anomaly text axis carries discriminative signal inside the low-rank subspace and cannot be naively orthogonalized.
(ii) \textbf{Limited Stage-2 effect on pure text maps:} applying Stage~2 alone (\texttt{Base+S2}) is near-neutral on MVTec ($81.69$ vs.\ $81.50$) and slightly negative on VisA ($74.77$ vs.\ $75.08$), while adding Stage~2 atop \texttt{S1-Res} adds only $+0.10$ (MVTec $88.51$) and $+0.21$\,pp (VisA $78.95$), reinforcing Stage~1 as the primary driver.

\begin{table}[!htbp]
\centering
\renewcommand{\arraystretch}{1.10}
\setlength{\tabcolsep}{4.6pt}
\caption{\textbf{Pure-text zero-shot VLM extension on official AnomalyCLIP (\texttt{ViT-L/14@336px}, official \texttt{epoch15} cross-dataset checkpoints, $K=4$ post-hoc calibration, $r=32$, seed 43, \%).} Zero visual memory bank and zero weight updates; \texttt{Base (Official)} is our unified 6-metric reproduction. Top block summarizes channel ($\Delta_{\mathrm{txt}} = q - t_{\mathrm{norm}}$) and spatial 9-D diagnostics.}
\label{tab:vlm_anomalyclip}
\resizebox{\linewidth}{!}{%
\begin{tabular}{@{}lccccccc@{\hspace{10pt}}ccccccc@{}}
\toprule
 & \multicolumn{7}{c}{\textbf{MVTec-AD (15 classes, VisA-trained ckpt)}} & \multicolumn{7}{c}{\textbf{VisA (12 classes, MVTec-trained ckpt)}} \\
\cmidrule(lr){2-8}\cmidrule(lr){9-15}
\rowcolor{panelbg}\textbf{Diagnostic} & \multicolumn{7}{c}{\scriptsize $r_{\mathrm{eff}}/D=0.004$, $E_{32}=95.9/95.4\%$, $r_{\mathrm{eff},9}=0.368$, TPR@5\%$=70.8/71.7/\mathbf{73.6}\%$} & \multicolumn{7}{c}{\scriptsize $r_{\mathrm{eff}}/D=0.004$, $E_{32}=95.6/95.3\%$, $r_{\mathrm{eff},9}=0.272$, TPR@5\%$=44.9/48.2/\mathbf{53.1}\%$} \\
\midrule
\textbf{Configuration} & \textbf{I-AUC} & \textbf{I-AP} & \textbf{I-$F_1$} & \textbf{P-AUC} & \textbf{P-$F_1$} & \textbf{PRO} & \textbf{Avg6} & \textbf{I-AUC} & \textbf{I-AP} & \textbf{I-$F_1$} & \textbf{P-AUC} & \textbf{P-$F_1$} & \textbf{PRO} & \textbf{Avg6} \\
\midrule
\texttt{Base (Official)} & 90.52 & 95.77 & 91.33 & 88.99 & 39.41 & 82.99 & 81.50 & 79.98 & 83.68 & 79.68 & 94.41 & 28.47 & 84.25 & 75.08 \\
\texttt{Base+S2} & 91.89 & 96.02 & 92.16 & 88.62 & 38.22 & 83.22 & 81.69 & 81.51 & 84.47 & 80.30 & 92.97 & 27.59 & 81.74 & 74.77 \\
\texttt{S1-Axis} & 93.61 & 97.40 & 93.37 & 82.43 & 39.64 & 80.20 & 81.11 & 80.57 & 83.45 & 79.43 & 83.96 & 27.47 & 67.95 & 70.47 \\
\texttt{S1-Axis+S2} & \textbf{94.35} & \textbf{97.65} & 94.23 & 87.13 & 39.55 & 83.62 & 82.76 & 81.70 & 84.55 & 80.08 & 89.13 & 26.97 & 75.19 & 72.94 \\
\rowcolor{deprinttint}\texttt{S1-Res} & 93.66 & 97.22 & 93.70 & 96.56 & \textbf{59.41} & \textbf{89.93} & 88.41 & 84.04 & 85.26 & 82.55 & \textbf{96.56} & \textbf{37.76} & \textbf{86.29} & 78.74 \\
\rowcolor{deprinttint}\texttt{S1-Res+S2} & 93.84 & 97.13 & \textbf{94.29} & \textbf{96.88} & 59.03 & 89.90 & \textbf{88.51} & \textbf{85.28} & \textbf{87.07} & \textbf{82.94} & 96.43 & 36.85 & 85.13 & \textbf{78.95} \\
\bottomrule
\end{tabular}}
\end{table}

\end{document}